\documentclass[11pt,a4paper]{article}
\usepackage[preprint]{acl}
\usepackage{times}
\usepackage{latexsym}

\usepackage[T1]{fontenc}

\usepackage{amssymb}
\usepackage{graphicx}
\usepackage{todonotes}
\usepackage{amsmath}
\usepackage{bm}
\usepackage{enumitem}
\usepackage{graphicx}
\usepackage{multirow}
\usepackage{cleveref}
\usepackage{tikz}
\usepackage[edges]{forest}
\usepackage{graphicx}
\usepackage{subcaption}
\usepackage{adjustbox}
\usepackage[utf8]{inputenc}
\usepackage{enumitem}
\usepackage{microtype}
\usepackage{enumitem}

\usepackage{pgfplots}
\definecolor{paired-light-blue}{RGB}{198, 219, 239}
\definecolor{paired-dark-blue}{RGB}{49, 130, 188}
\definecolor{paired-light-orange}{RGB}{251, 208, 162}
\definecolor{paired-dark-orange}{RGB}{230, 85, 12}
\definecolor{paired-light-green}{RGB}{149, 203, 211}
\definecolor{paired-dark-green}{RGB}{56, 177, 195}
\definecolor{paired-light-purple}{RGB}{218, 218, 235}
\definecolor{paired-dark-purple}{RGB}{117, 107, 176}
\definecolor{paired-light-gray}{RGB}{217, 217, 217}
\definecolor{paired-dark-gray}{RGB}{99, 99, 99}
\definecolor{paired-light-pink}{RGB}{222, 158, 214}
\definecolor{paired-dark-pink}{RGB}{123, 65, 115}
\definecolor{paired-light-red}{RGB}{238, 173, 152}
\definecolor{paired-dark-red}{RGB}{241, 131, 95}
\definecolor{paired-light-yellow}{RGB}{231, 204, 149}
\definecolor{paired-dark-yellow}{RGB}{141, 109, 49}
\definecolor{light-green}{RGB}{118, 207, 180}
\definecolor{raspberry}{RGB}{228, 24, 99}
\definecolor{pastelblue}{HTML}{C9DAEA}
\definecolor{lightteal}{HTML}{B4D1CD}
\definecolor{softgray}{HTML}{BEBEBE}

\usepackage[utf8]{inputenc}
\usepackage{booktabs}  
\usepackage{subcaption}

\usepackage{url}

\title{Isolating Culture Neurons in Multilingual Large Language Models}

\author{Danial Namazifard \\
  University of Tehran\\
  {\tt namazifard@ut.ac.ir} \\\And
  Lukas Galke Poech\\
  University of Southern Denmark\\
  {\tt galke@imada.sdu.dk} }

\date{}

\begin{document}

\maketitle
\begin{abstract}
    Language and culture are deeply intertwined, yet it has been unclear how and where multilingual large language models encode culture.
    Here, we build on an established methodology for identifying language-specific neurons to localize and isolate culture-specific neurons, carefully disentangling their overlap and interaction with language-specific neurons. 
    To facilitate our experiments, we introduce MUREL, a curated dataset of 85.2 million tokens spanning six different cultures.
    Our localization and intervention experiments show that LLMs encode different cultures in distinct neuron populations, predominantly in upper layers, and that these culture neurons can be modulated largely independently of language-specific neurons or those specific to other cultures.
    These findings suggest that cultural knowledge and propensities in multilingual language models can be selectively isolated and edited, with implications for fairness, inclusivity, and alignment. Code and data are available at \url{https://github.com/namazifard/Culture_Neurons}.
\end{abstract}

\section{Introduction}
Cultural context underpins human communication, shaping interpretations, values, and worldviews that go beyond linguistic surface forms~\cite{kramsch2014language}.
For example, opinions on morality, authority, and gender roles can vary dramatically across cultural groups, even when expressed in the same language. 
Recent advances in multilingual large language models (LLMs)~\cite{team2025gemma} have drawn increased attention on their cultural propensities. Understanding and potentially controlling the cultural propensities of language models is crucial to ensure cultural fairness, inclusivity, and alignment~\cite{liu2025culturally}.

\begin{figure}[!th]
    \centering
    \includegraphics[width=1.01\linewidth]{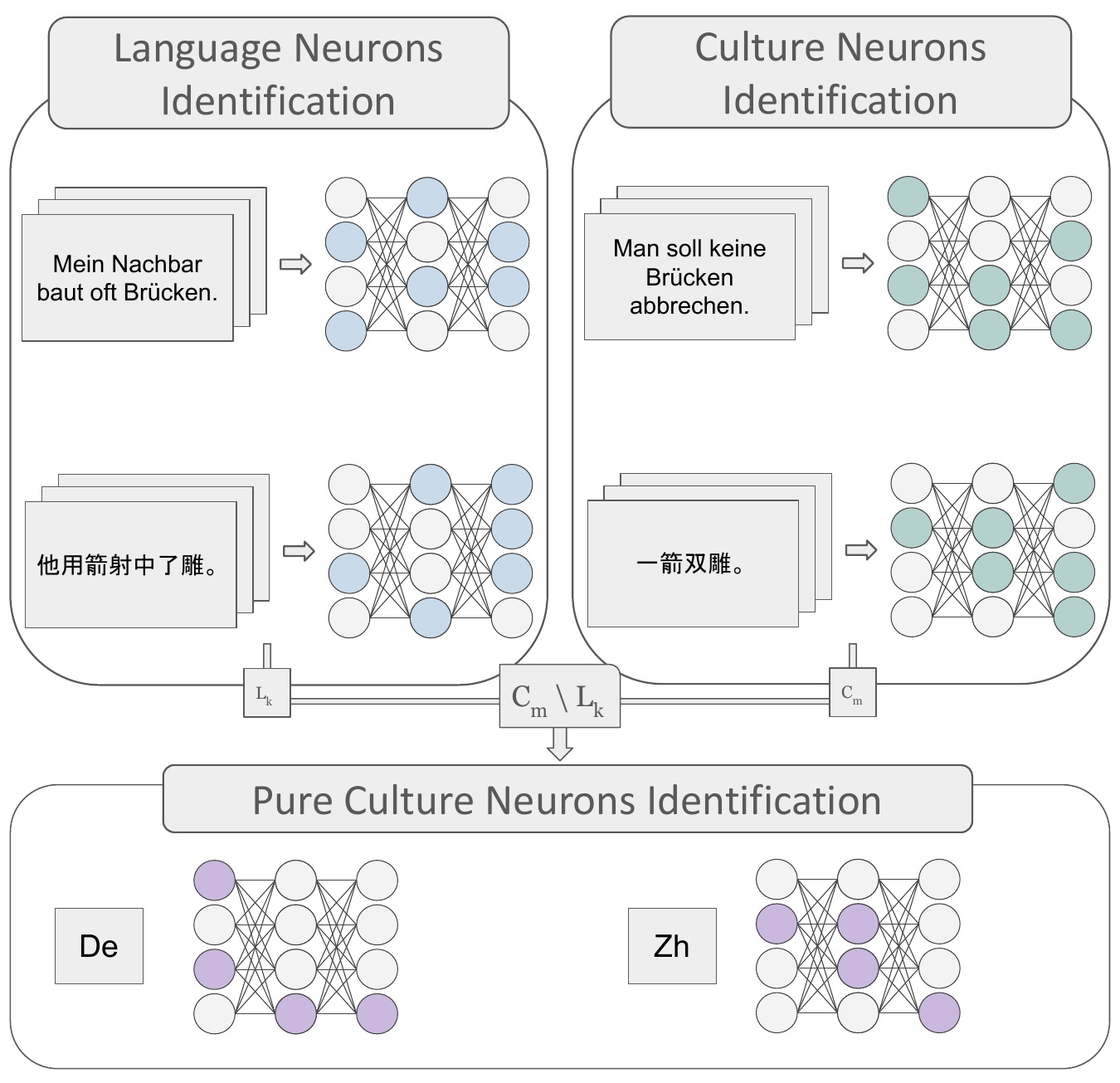}
    \caption{Overview of our methodology for identifying pure culture-specific neurons in language models. We first identify language-specific neurons ($\mathbb{L}_k$) using literal, language-focused sentences (left), and culture-specific neurons ($\mathbb{C}_m$) using culturally salient phrases (right). By subtracting the language-specific neuron set from the culture-specific neuron set, we obtain pure culture-specific neurons ($\mathbb{C}_m \setminus \mathbb{L}_k$), which encode culture independently of language (bottom).}
    \label{fig:overview}
\end{figure}

Here, we set out to study model internals governing such cultural propensities. Can we localize a set of neurons that drives cultural propensities? Is this set of neurons separate from the language associated with that culture? Can we intervene on the model internals to modulate cultural propensities without any training?

Despite advances in neuron localization and editing techniques~\cite{dai-etal-2022-knowledge, hou-etal-2023-towards, li2023inference}, isolating \emph{pure culture-specific neurons}, i.e., those that drive culture but not language, remains particularly challenging, given their inherent entanglement~\cite{liu2025culturally,kramsch2014language} and even data on non-linguistic elements of culture needs to be encoded linguistically when fed into a language model.

Prior work on localization of language-specific neurons has revealed that different languages are encoded in different areas of the model~\cite{tang-etal-2024-language, zhao2024large}. However, these methods are insufficient for identifying and isolating culture-specific neurons, due to the expected entanglement with language and the lack of suitable datasets.

To address these challenges, we develop a methodology for identifying both culture-specific and pure culture-specific neurons, by which we mean those neurons that are specific to a culture but not specific to the language associated with that culture. To facilitate this methodology, we compile a dataset covering different linguistic and non-linguistic cultural elements.

Our results indicate that, despite the inherent entanglement of language and culture, it is possible to identify neuron populations more strongly associated with culture than language. 
We find that cultural representations are localized in regions often distinct from language-encoding neurons, with different cultures occupying separate neural populations. This allows selective modulation of one culture's representations largely independently of other cultures.

In sum, the contributions of this work are:

\begin{itemize}[noitemsep]
    \item We introduce a methodology for identifying (pure) culture neurons in LLMs (\S\ref{sec:methodology}).
    \item We present MUREL, a dataset of 85.2 million tokens covering six cultures (\S\ref{sec:datasets}).
    \item We conduct experiments  showing that culture-specific neurons are largely separable from language-specific neurons (\S\ref{sec:exp-setup} and \ref{sec:neuron_idf}).
    \item We show that culture neurons can be selectively modulated (\S\ref{subsec:ablation_intervention}).
\end{itemize}
\section{Related Work}

% \subsection{Evaluating Cultural Competence}
Recent work on multilingual language models has primarily focused on linguistic capabilities~\citep{liang2022holistic, srivastava2023beyond, ahuja-etal-2023-mega}. More recently, research has shifted toward assessing cultural competence, including culturally salient elements such as norms \citep{ziems-etal-2023-normbank}, values \citep{moore-etal-2024-large}, and worldviews \citep{mushtaq2025worldview}. Some studies have examined cultural knowledge within a monocultural setting \citep{muller-eberstein-etal-2025-dakultur}, while a growing body of work investigates multicultural evaluations, exploring cultural phenomena across languages and societies~\citep{yin-etal-2022-geomlama, fung-etal-2023-normsage, huang2025mceval}. Recent efforts have extended this research to vision-language models, evaluating cultural understanding within monocultural and multicultural contexts \citep{alwajih-etal-2024-peacock,romero2024cvqa, vayani2025all, nayak-etal-2024-benchmarking}.

Despite these advances, existing evaluations are predominantly behavioral, leaving open critical questions about how cultural knowledge is internally encoded in multilingual models.

% \subsection{Mechanistic Interpretability}
% \paragraph{}
In parallel, mechanistic interpretability research has sought to uncover how LLMs encode information at the neuron level. These studies have successfully identified neurons or populations of neurons corresponding to specific capabilities, such as knowledge storage \citep{dai-etal-2022-knowledge}, safety alignment \citep{chen2024finding}, and confidence estimation \citep{stolfo2024confidence}. Recent work has further explored specialized neurons in multilingual models, discovering neurons encoding language identity or linguistic features~\citep{tang-etal-2024-language, zhao2024large, kojima-etal-2024-multilingual}, with similar findings for vision-language models~\citep{huo-etal-2024-mmneuron}.

However, while these advances have deepened our understanding of language models' representations of linguistic and semantic information, they have not addressed the neural encoding of cultural knowledge independent of linguistic identity. This paper addresses this gap by introducing a systematic methodology for studying culture-specific neurons, providing new insights into the cultural representations in language models.
\section{Identifying Culture Neurons}
\label{sec:methodology}
Our goal is to identify and isolate \emph{culture-specific neurons} within multilingual LLMs. To disentangle language and culture, we develop a systematic approach to distinguish neuron populations that respond specifically to cultural inputs, independently of linguistic features. This requires first identifying language-specific and culture-specific neurons, and then applying set operations to isolate pure culture neurons, as described below.

\subsection{Background: Identification of Language-Specific Neurons}
\label{sec:method-lape}

We first locate \emph{language-specific neurons} as the basis for disentangling linguistic and cultural factors. We adopt the language activation probability entropy (LAPE) method~\citep{tang-etal-2024-language}, which effectively detects language-localized regions in multilingual LLMs. We briefly recapitulate their approach, as it forms the foundation for our identification of culture-specific neurons.

Modern LLMs are built on autoregressive transformer architectures~\citep{Vaswani-etal-2017-attention} with multi-head self-attention (MHA) and feed-forward networks (FFNs). 
Let $\tilde{\bm{h}}^{\ell}$ denote the output of the MHA module in the $\ell$-th layer, computed using the previous layer's hidden states and trainable parameters, and $\operatorname{act\_fn}(\cdot)$ denotes the activation function. The FFN output $\bm{h}^{\ell} \in \mathbb{R}^{d_1}$ in a GLU variant is:
\begin{equation*}
    \bm{h}^{\ell} = (\text{act\_fn}(\tilde{\bm{h}}^{\ell}\mathbf{W}^{(\ell)}_1) \otimes \tilde{\bm{h}}^{\ell}\mathbf{W}^{(\ell)}_3\big) \cdot \mathbf{W}^{(\ell)}_2,
\end{equation*}
where
$\mathbf{W}^{(\ell)}_1, \mathbf{W}^{(\ell)}_3 \in \mathbb{R}^{d_1 \times d_2}$ and $\mathbf{W}^{(\ell)}_2 \in \mathbb{R}^{d_2 \times d_1}$ are learnable parameters. In LAPE, a \textit{neuron} is defined as the linear transformation of a single column in $\mathbf{W}^{(\ell)}_1$ followed by the application of the non-linear activation function. Thus, each FFN module contains $d_2$ neurons.

LAPE identifies neurons with systematically different activation probabilities across languages. For neuron $j$ in layer $\ell$ and language $k$:
\begin{equation*}
    p^k_{\ell,j} = \mathbb{E}\left[ \mathbb{I}(\text{act\_fn}(\tilde{\bm{h}}^{\ell}\mathbf{W}^{(\ell)}_1)_j > 0) ~\middle|~ \text{language } k \right],
\end{equation*}
where $\mathbb{I}(\cdot)$ is the indicator function. The activation probability is empirically estimated by the likelihood that the neuron’s activation value exceeds zero. The probabilities across all languages $\mathcal{L}$ yield a distribution $\bm{p}_{\ell,j} = (p^1_{\ell,j}, ..., p^k_{\ell,j}, ..., p^l_{\ell,j})$, which is normalized:
$
    p'^k_{\ell,j} = \frac{p^k_{\ell,j}}{\sum_{k' \in \mathcal{L}} p^{k'}_{\ell,j}}
    $
The entropy of this distribution is:
$
    \text{LAPE}_{\ell,j} = -\sum_{k \in \mathcal{L}} p'^k_{\ell,j} \log p'^k_{\ell,j}
    $
Neurons with low LAPE are highly language-specific. We define language-specific neurons as those in the bottom 1\% of LAPE, requiring that at least one language has an activation probability above a specified threshold.
In practice, we follow~\citet{tang-etal-2024-language} by using balanced corpora per language and computing LAPE scores, yielding a sparse set $\mathbb{L}_k$ for each language $k$.

\subsection{Identification of Culture-Specific Neurons}
\label{sec:method-cape}

Similarly, we define \emph{Culture Activation Probability Entropy} (CAPE) by evaluating activation probabilities over culturally distinct inputs. For culture $m$, the activation probability for neuron $j$ in layer $\ell$ is defined as:
\begin{equation*}
    q^m_{\ell,j} = \mathbb{E}\left[ \mathbb{I}(\text{act\_fn}(\tilde{\bm{h}}^{\ell}\mathbf{W}^{(\ell)}_1)_j > 0) ~\middle|~ \text{culture } m \right],
\end{equation*}
We normalize and compute entropy as with LAPE, and define \emph{culture-specific neurons} $\mathbb{C}_m$ as those with CAPE below the threshold $\tau_{\text{cult}}$.
\begin{equation*}
    \mathbb{C}_m = \{ v \in \mathbb{N} \mid \text{CAPE}(v) \leq \tau_{\text{cult}} \}
\end{equation*}

\subsection{Disentangling Culture from Language}
\label{subsec:Disentangling_Culture_Language}
To disentangle language and culture, we apply set operations at the neuron level.
Let $\mathbb{N}$ denote the set of all FFN neurons in a given model.  
For each language $k$, we identify a subset $\mathbb{L}_k \subset \mathbb{N}$ of \emph{language-specific neurons} using the LAPE method (\S\ref{sec:method-lape}).  
Similarly, for each culture $m$, we define the set $\mathbb{C}_m \subset \mathbb{N}$ of \emph{culture-specific neurons} (\S\ref{sec:method-cape}).

We assume that there is some overlap between language and culture neurons. To isolate \emph{pure culture-specific neurons} for culture $m$ and its associated language $k$, we define:
$
    \mathbb{P}_m = \mathbb{C}_m \setminus \mathbb{L}_k
    $
as the set of neurons specific to culture $m$ that are not language-specific. 
Some neurons may respond to both language and culture, which we define as \emph{compound language-and-culture neurons} $\mathbb{L}_k \cap \mathbb{C}_m$.
% Finally, we also consider the set of neurons that are active for all tested cultures $\mathbb{G} = \bigcap_m \mathbb{C}_m$.
%
This framework partitions neuron space into language, culture, compound, and generic components.

\subsection{Interventions}
We assess the functional roles of these neuron subsets by systematically deactivating (zeroing out) their activations during inference. 

\textbf{Deactivating neuron subpopulations:} Given a set of neurons $\mathbb{X}$ -- where $\mathbb{X} \in \{\mathbb{L}_k,\, \mathbb{C}_m,\, \mathbb{P}_m,\, \mathbb{L}_k \cap \mathbb{C}_m\}$, representing language-specific, culture-specific, pure-culture, and compound neurons, respectively -- we set all activations in $\mathbb{X}$ to zero. 

\textbf{Random neuron ablation:} As a control, we select and deactivate a size-matched amount of random neurons to ensure observed effects are not due to neuron count alone.
\section{MUREL: A Multicultural Resource for Evaluating Language Models}\label{sec:datasets}

To support our neuron analysis, we introduce \textbf{MUREL} (\textbf{MU}lticultural \textbf{R}esource for \textbf{E}valuating \textbf{L}anguage Models), a comprehensive dataset collection spanning culturally diverse text resources. MUREL is constructed from public sources and systematically organized according to the taxonomy proposed by \citet{liu2025culturally}, enabling broad coverage of ideational, linguistic, and social dimensions for targeted analysis of culture-specific and linguistic phenomena.
In total, MUREL comprises 69 datasets spanning six cultural groups, containing an average of 14.2 million tokens per culture (see Appendix~\ref{app:A} for detailed statistics).

\subsection{Dataset Organization}
We categorize the datasets into three primary branches, as defined by \citet{liu2025culturally}: (i) \textbf{Ideational Elements}, covering abstract cultural concepts and knowledge; (ii) \textbf{Linguistic Elements}, focusing on intra-linguistic variations and communicative styles; and (iii) \textbf{Social Elements}, encompassing factors related to human interactions and demographic attributes.
An overview of the datasets along these dimensions is described below. 

\paragraph{Ideational Elements} comprise concepts, knowledge, values, norms, morals, and artifacts.
\underline{Concepts}
are salient, lexicalized ideas representing either culturally unique objects or figurative expressions, for which we use data on metaphors~\cite{kabra-etal-2023-multi}, proverbs and sayings~\cite{liu-etal-2024-multilingual}, idioms~\cite{stap-etal-2024-fine, khoshtab-etal-2025-comparative}, and ironies~\cite{casola-etal-2024-multipico}.
\underline{Knowledge}:
Culture-specific factual and common-sense information
is covered through cultural probing datasets \cite{bhatt-diaz-2024-extrinsic}, multiple-choice QA benchmarks \cite{wang-etal-2024-seaeval}, and knowledge bases capturing cultural knowledge~\cite{koto-etal-2024-indoculture}. 

\underline{Values}
represent beliefs and behavioral standards prioritized differently across cultural groups. To capture these, we combine established resources such as the Pew Global Attitudes Survey (PEW)\footnote{\url{https://www.pewresearch.org/}}, World Values Survey (WVS)\footnote{\url{https://www.worldvaluessurvey.org/}}, Political Compass Test (PCT)\footnote{\url{https://www.politicalcompass.org/}}, and Hofstede's Cultural Dimensions~\cite{hofstede1984culture}.  Additionally, we consider recent NLP datasets specifically developed to assess the alignment and manifestation of cultural values in large language models \cite{cao-etal-2023-assessing, pistilli2024civics, lee-etal-2024-exploring-cross}.

\underline{Norms and Morals}
are sets of culture-dependent principles governing acceptable behaviors and judgments. To cover this area, we utilize existing norm banks~\cite{dwivedi-etal-2023-eticor, ch-wang-etal-2023-sociocultural, fung-etal-2023-normsage}. Additionally, we incorporate datasets that employ direct querying of language models on ethical and normative issues~\cite{yuan-etal-2024-measuring, yu-etal-2024-cmoraleval}.

\underline{Artifacts}
include culturally significant products of human creativity such as literature, poetry, music, films, and memes. Our compilation incorporates datasets covering literary texts, fairy tales, and poetry, designed explicitly for cultural analysis and cross-cultural adaptation~\cite{yang-etal-2019-generating, chakrabarty-etal-2021-dont, schmidt-etal-2021-fairynet}.

\paragraph{Linguistic Elements} cover dialects, styles, registers, and genres. 
\underline{Dialects}
are systematic linguistic variants influenced by regional, national, or sociocultural factors.

To encompass dialectal diversity, our compilation integrates datasets designed for dialect identification and analysis \cite{malmasi-zampieri-2017-german, ciobanu-etal-2018-german} as well as resources focusing on translations between dialects and standard languages \cite{pluss-etal-2023-stt4sg, kuparinen-etal-2023-dialect}.
\underline{Styles, Registers, Genres}
include linguistic variations shaped by situational context, communicative goals, and societal norms. Our compilation incorporates datasets designed to evaluate style and register in NLP tasks, focusing on aspects such as formality~\cite{nadejde-etal-2022-cocoa}, politeness~\cite{srinivasan-choi-2022-tydip, havaldar-etal-2023-comparing}, slang~\cite{sun-xu-2022-tracing}, and genre-specific language, including news reporting and storytelling.

\paragraph{Social Elements} cover relationships, context, communicative goals, and demographics.
\underline{Relationship}
addresses how communication varies according to interpersonal and societal connections, such as family roles or social hierarchies. Our collection includes datasets that explicitly account for culture-specific relationship terms and interaction dynamics~\cite{abs-2402-11178}, ensuring nuanced modeling of communication styles sensitive to relationship contexts.

\underline{Context} refers to the linguistic and extra-linguistic settings shaping communication, such as situational, historical, or non-verbal cues. To comprehensively address contextual variation, our dataset compilation includes resources emphasizing both textual contexts and broader frames of reference~\cite{hovy-etal-2020-sound, chakrabarty-etal-2022-rocket, zhan2023social, ziems-etal-2023-normbank}.

\underline{Communicative Goals}
cover culturally distinct purposes behind language use, including indirect versus direct communication styles in refusal, requests, and apologies. We incorporate resources tailored to evaluating these pragmatic variations, supporting tasks that require understanding shaped communicative intents and their linguistic expressions~\cite{emelin-etal-2021-moral, li-etal-2023-normdial, abs-2402-11178}.

\underline{Demographics}
reflect characteristics of individuals and groups, such as age, income, educational level, or ethnicity, which influence communication patterns. Our dataset selection includes demographic-focused datasets that facilitate exploration of how sociodemographic attributes impact linguistic usage and perception~\cite{voigt-etal-2018-rtgender, hovy-etal-2020-sound, santy-etal-2023-nlpositionality}.

\subsection{Language Selection}
For our study, we selected six typologically and geographically diverse languages: English (\emph{en}), German (\emph{de}), Danish (\emph{da}), Chinese (\emph{zh}), Russian (\emph{ru}), and Persian (\emph{fa}). The selection was guided by three criteria: (a) \textbf{geographical diversity}, covering Western Europe, East Asia, Eastern Europe, and the Middle East; (b) \textbf{linguistic typology}, including the Germanic, Slavic, Indo-Iranian, and Sino-Tibetan language families; and (c) \textbf{resource availability}, spanning both high-resource (e.g., English) and lower-resource (e.g., Danish) languages. This diversity enables us to assess the robustness of neuron detection methods across a broad spectrum of linguistic and cultural contexts, enhancing the generalizability of our findings beyond any single language, culture, or region.

\begin{figure*}[t]
    \centering
    \begin{subfigure}[b]{0.47\textwidth}
        \includegraphics[width=\textwidth]{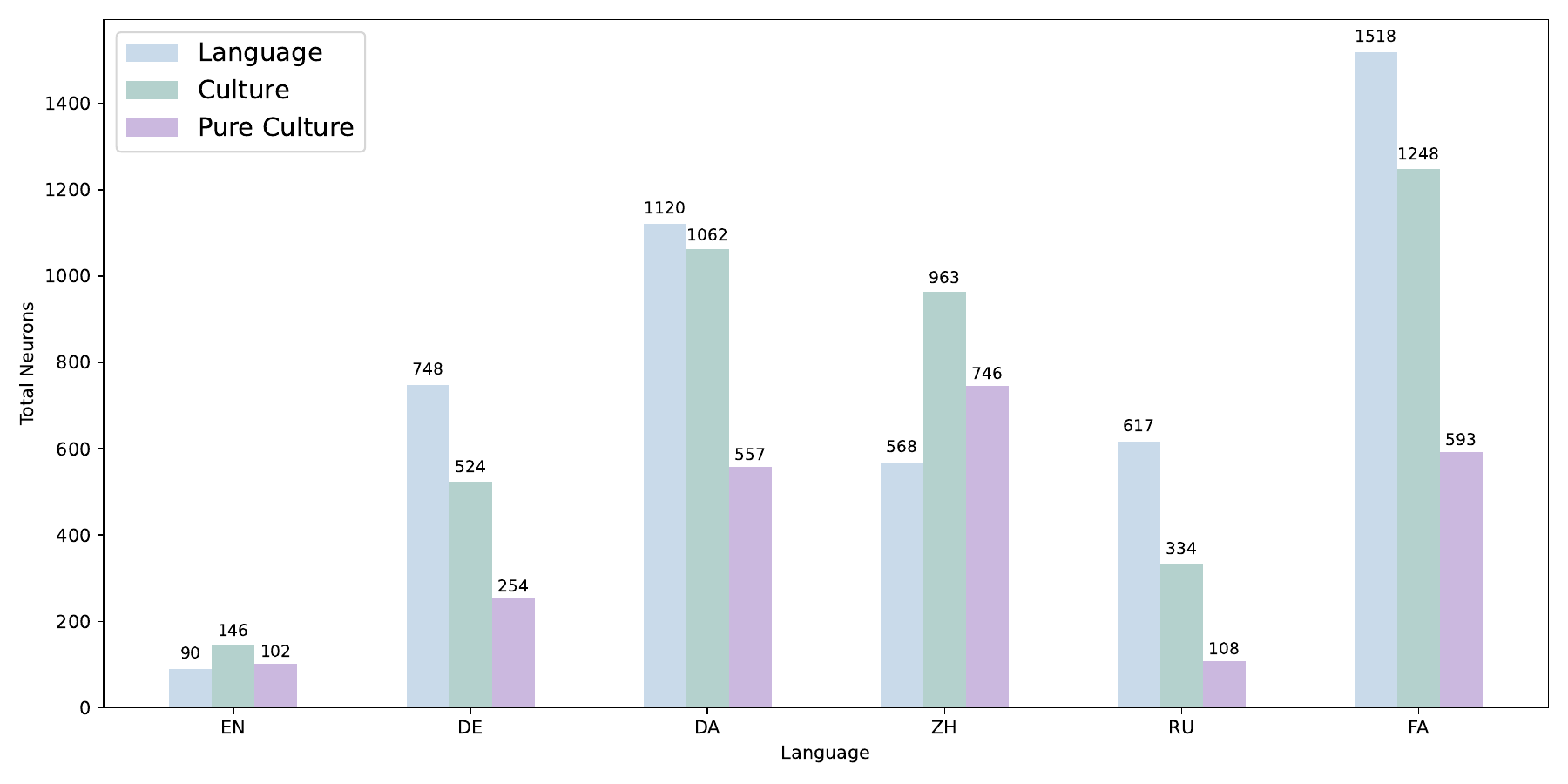}
        \caption{Llama-2-7b}
    \end{subfigure}
    \begin{subfigure}[b]{0.47\textwidth}
        \includegraphics[width=\textwidth]{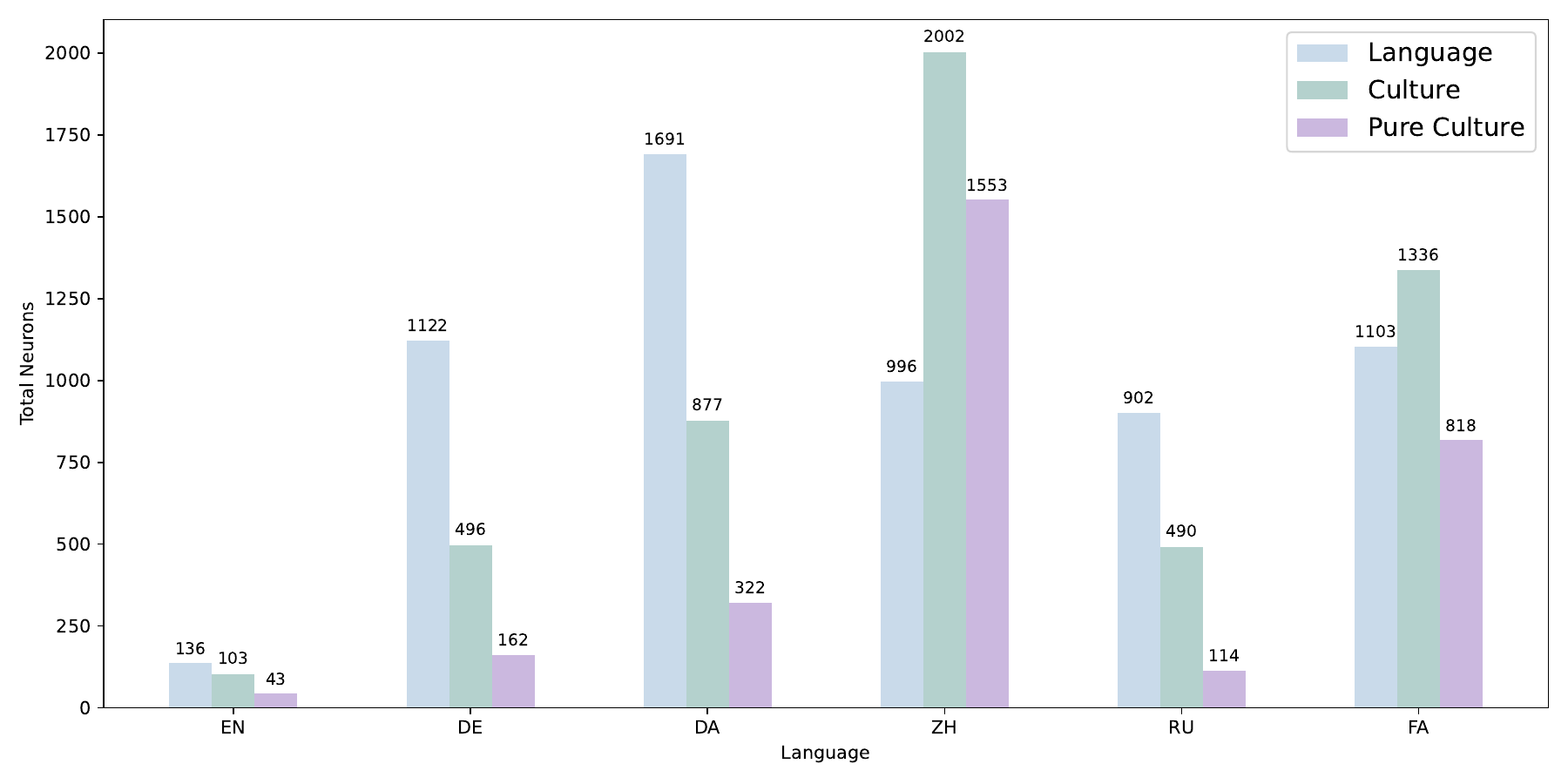}
        \caption{Llama-3.1-8b}
    \end{subfigure}
    \\
    \begin{subfigure}[b]{0.47\textwidth}
        \includegraphics[width=\textwidth]{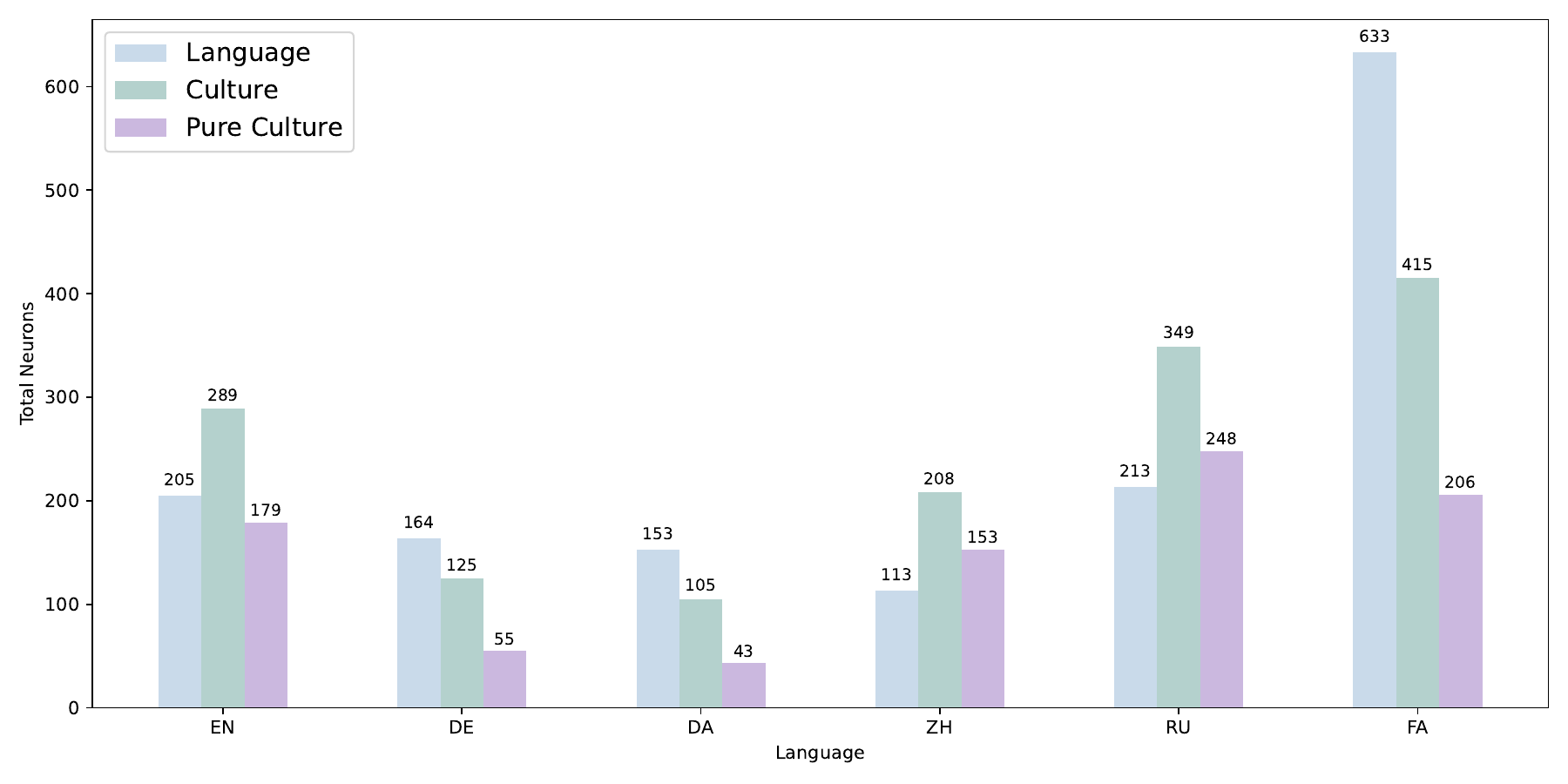}
        \caption{Qwen2.5-7b}
    \end{subfigure}
    \begin{subfigure}[b]{0.47\textwidth}
        \includegraphics[width=\textwidth]{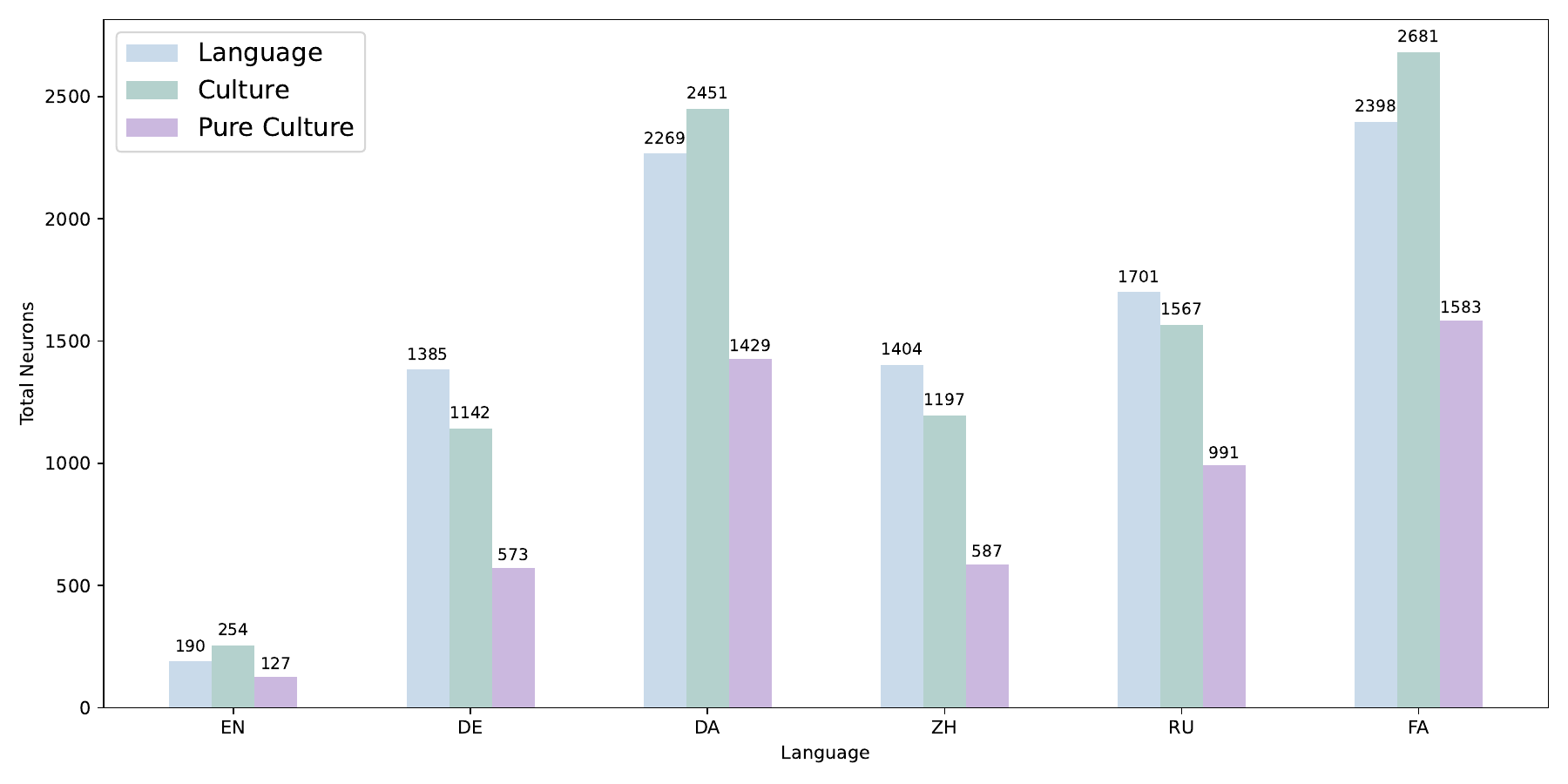}
        \caption{Gemma-3-12b}
    \end{subfigure}
    \caption{Total language, culture, and pure culture neurons per language and model. The total number of identified neurons is 3,523 for Llama-2-7b, 4,588 for Llama-3.1-8b, 1,004 for Qwen2.5-7b, and 7,373 for Gemma-3-12b.}
    \label{fig:grouped_lang_bar_all_models}
\end{figure*}

\subsection{Dataset Preparation}
To ensure that each dataset was suitable for detecting culture-specific neurons, we performed targeted adaptation and reformatting for several datasets. While some datasets could be directly integrated in their original format, others required modification to better align with our experimental setup and enabling finer-grained cultural analysis.

For example, we transformed original World Values Survey (WVS) probes into textual statements that explicitly encode cultural nuances. Original survey items, such as, ``Familie ist [MASK] in meinem Leben'' with possible responses ``wichtig'' or ``unwichtig'' were converted into complete statements (e.g., \textit{Familie ist wichtig in meinem Leben}).

\noindent Such transformations allow us to treat each response as an independent cultural assertion and standardize inputs while preserving cultural information.
\section{Experimental Setup}
\label{sec:exp-setup}

Our goal is to systematically identify \textbf{pure culture-specific neurons} that respond specifically to cultural information, independent of language, within multilingual LLMs.
We proceed as follows:

We first apply the language activation probability entropy (LAPE) method (\S\ref{sec:method-lape}) to Wikipedia corpora\footnote{\url{https://huggingface.co/datasets/wikimedia/wikipedia}}. For each language $k$, we use 100 million tokens to robustly capture neuron activation patterns across diverse linguistic contexts, following established methodology~\citep{tang-etal-2024-language}.

Next, we apply the culture activation probability entropy (CAPE) method (\S\ref{sec:method-cape}) to our MUREL dataset, using 10 million tokens per culture $m$.
Note that for each culture $m$, CAPE is computed over the \emph{union} of texts sampled from all three MUREL branches, i.e., the individual branches are \emph{not} considered as separate CAPE targets.
For evaluation, we use a separate, balanced held-out set of 100,000 tokens per culture, ensuring reliable measurement of neuron specificity.

Following prior work~\citep{tang-etal-2024-language},
we select the lowest 1\% of neurons by entropy as language- and culture-specific neurons for LAPE and CAPE, respectively. We use 1\% for sparsity and comparability. Prior studies found stable trends across cutoffs ranging from 1 to 10\%, and our pre-experiments were consistent with these findings. 

To disentangle the effects of language and culture, we categorize identified neurons as (1) \emph{Pure culture-specific neurons}: neurons that respond strongly to culture $m$ but are not language-specific; and (2) \emph{Compound language-and-culture neurons}, which respond to both language $k$ and culture $m$.

We conduct our experiments using four transformer-based pretrained language models, including \texttt{Llama-2-7b}~\citep{touvron2023llama}, \texttt{Llama-3.1-8b}~\citep{grattafiori2024llama}, \texttt{Qwen2.5-7b}~\citep{qwen2025qwen25technicalreport}, and \texttt{Gemma-3-12b}~\citep{team2025gemma}.

All models except Llama-2 are multilingual; Llama-2 is included as a monolingual baseline to test how our methodology generalizes beyond multilingual settings.
Additional details for each model are provided in Appendix~\ref{app:B}.
\section{Results}
\label{sec:results}
We first report neuron identification and distribution, then run intervention experiments to assess functional roles.

\subsection{Neuron Identification and Distribution}
\label{sec:neuron_idf}

\paragraph{Neuron Counts and Distributions}
Language- and culture-specific neurons were selected from all FFN layers based on the lowest activation entropy values.
Figure~\ref{fig:grouped_lang_bar_all_models} shows the number of language-specific, culture-specific, and pure culture-specific neurons identified across the four evaluated models for each tested language and culture.

Several notable patterns emerge:
First, the degree of neuron specialization varies not only by model but also across languages and cultures, reflecting distinct representational demands.

Second, lower-resource languages such as Persian and Danish show higher counts of both language-specific and culture-specific neurons compared to resource-rich languages like English. This aligns with prior work~\citep{tang-etal-2024-language}, suggesting that multilingual models allocate more representational capacity to underrepresented languages to capture richer linguistic and cultural nuances.

Third, although we explicitly target exactly 1\% of the neurons within FFN layers for both language- and culture-specific sets, we observe a slight discrepancy in the total neuron counts. This minor discrepancy arises naturally because some neurons simultaneously encode multiple languages or cultures, resulting in overlapping neuron sets.

Crucially, a substantial proportion (on average 56.7\%) of culture-specific neurons are categorized as \emph{pure culture-specific}, indicating they encode cultural representations largely independent of linguistic identity. This suggests that much of the cultural information within multilingual LLMs is neurally localized to specific populations of neurons that are, to a considerable degree, separable from language processing.

\begin{figure}[t]
    \centering
    \begin{subfigure}[b]{0.99\linewidth}
        \centering
        \includegraphics[width=\linewidth]{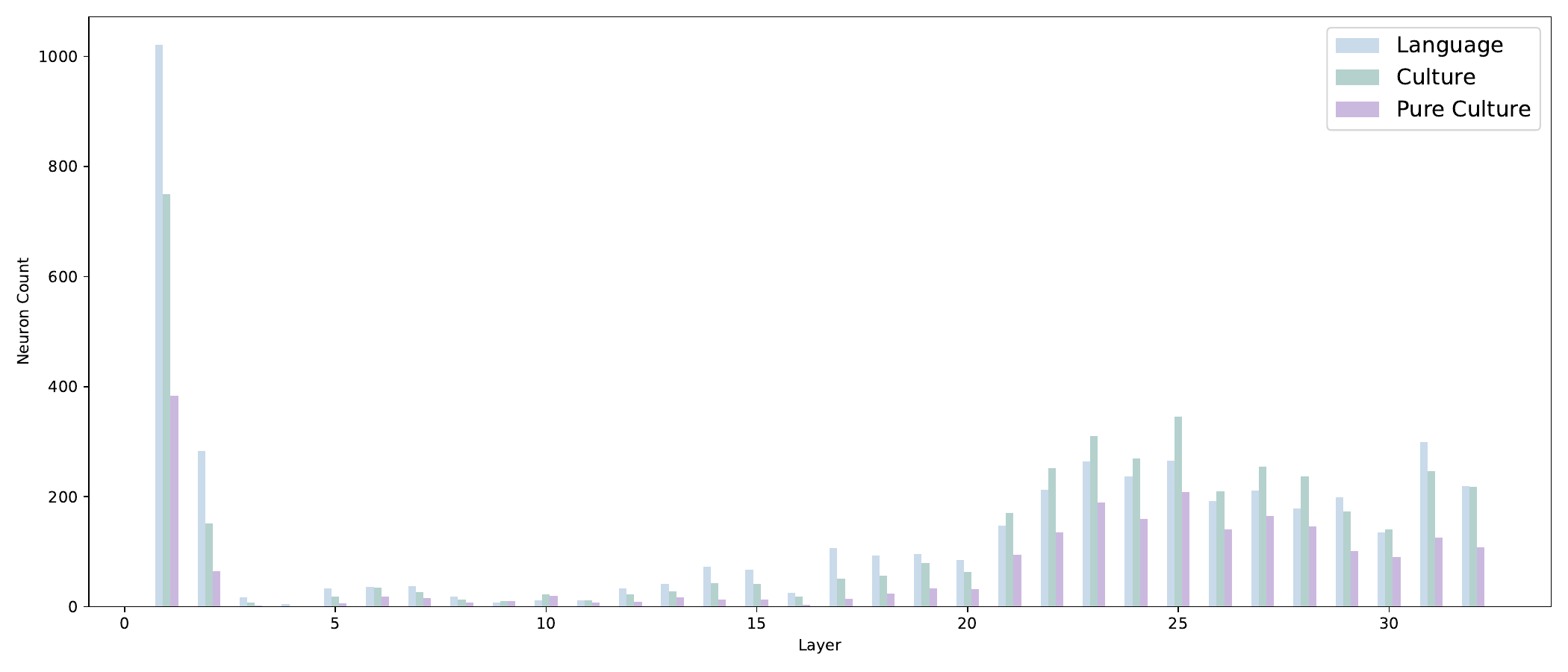}
        \caption{Llama-2-7b}
        \label{fig:layerwise_bar_Llama-2-7b}
    \end{subfigure}
    \vspace{1mm}
    \begin{subfigure}[b]{0.99\linewidth}
        \centering
        \includegraphics[width=\linewidth]{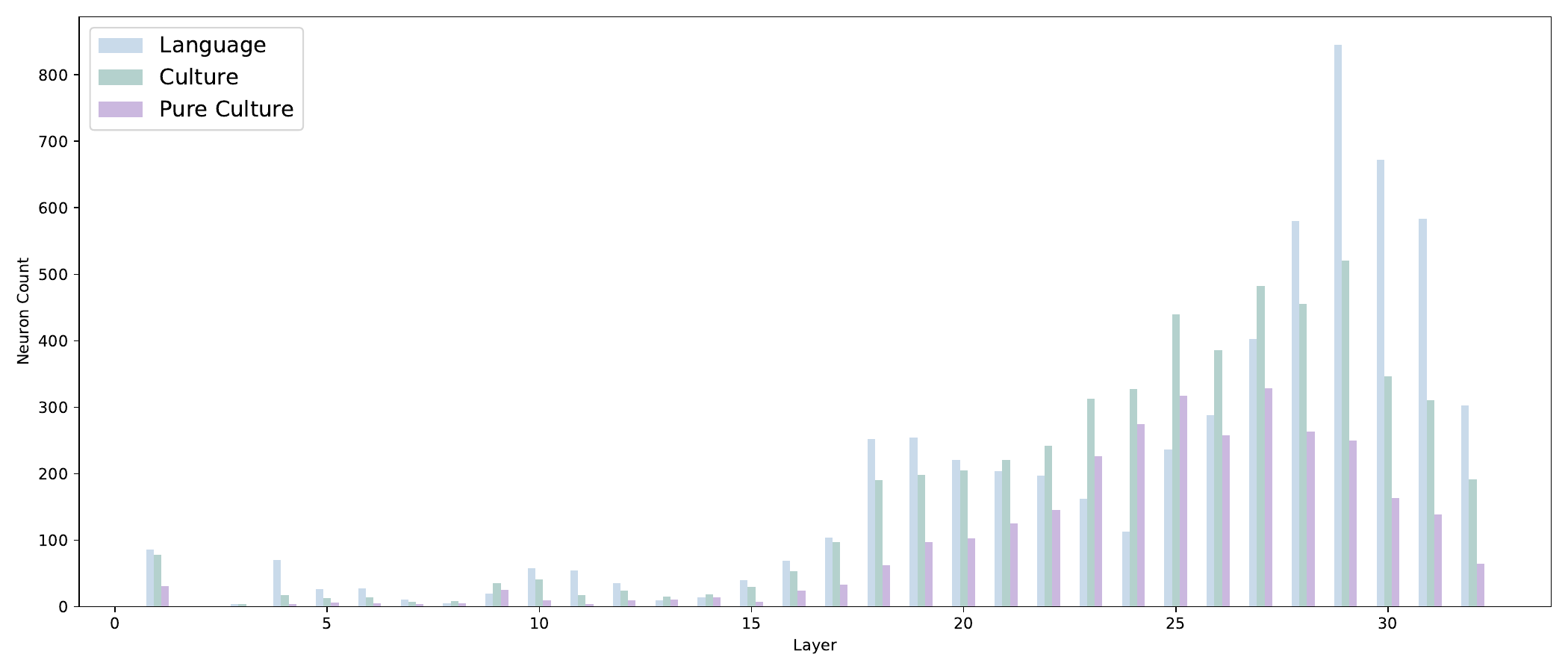}
        \caption{Llama-3.1-8b}
        \label{fig:layerwise_bar_Llama-3.1-8b}
    \end{subfigure}
    \caption{Layer-wise distribution of language-, culture-, and pure culture-specific neurons for models. Layer-wise distribution per language is shown in Figure~\ref{fig:all_languages_layerwise_line_both}.}\label{fig:layerwise_bar_models}
\end{figure}

\paragraph{Neuron Distribution across Layers.}
We next examine how neuron types are distributed across model layers. As shown in Figure~\ref{fig:layerwise_bar_models}, models tend to concentrate language- and culture-specific neurons in the upper layers. In Llama-2-7b, a monolingual model, we observe a secondary peak in the bottom layers, resulting in a bimodal distribution consistent with previous findings~\citep{tang-etal-2024-language, zhao2024large}. This may reflect a dual specialization, with early layers capturing lower-level linguistic patterns and top layers encoding higher-level semantics. By contrast, multilingual models (e.g. Llama-3.1-8b) show a more pronounced concentration of both neuron types exclusively in the upper layers, suggesting a more hierarchical organization of semantic information. Notably, pure culture-specific neurons follow a similar pattern but are comparatively sparser across layers.

\subsection{Intervention Experiments}
\label{subsec:ablation_intervention}

\begin{figure}[t!]
\centering
\begin{subfigure}{0.49\linewidth}
\includegraphics[width=\linewidth]{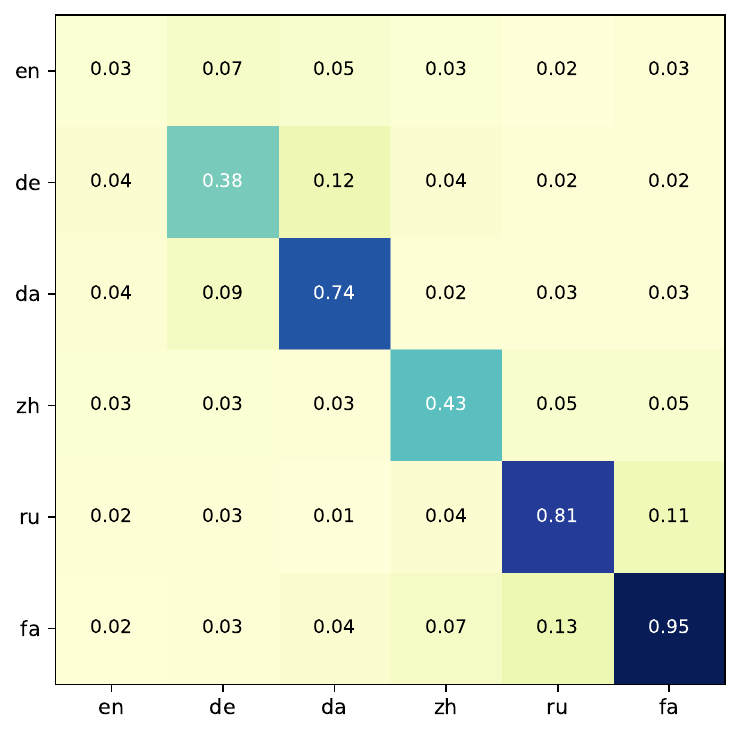}
\caption{Culture}
\end{subfigure}
\begin{subfigure}{0.49\linewidth}
\includegraphics[width=\linewidth]{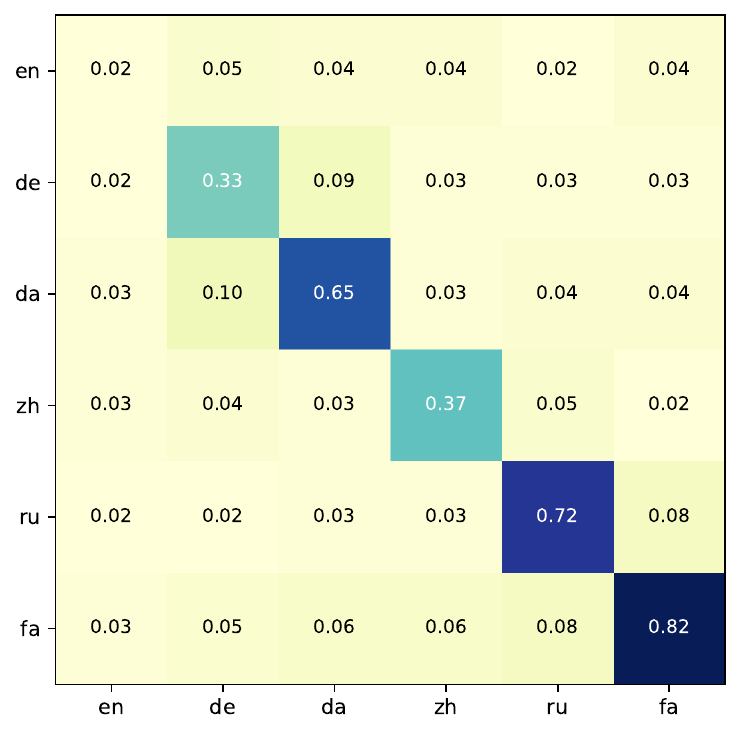}
\caption{Pure Culture}
\end{subfigure}
\begin{subfigure}{0.49\linewidth}
\includegraphics[width=\linewidth]{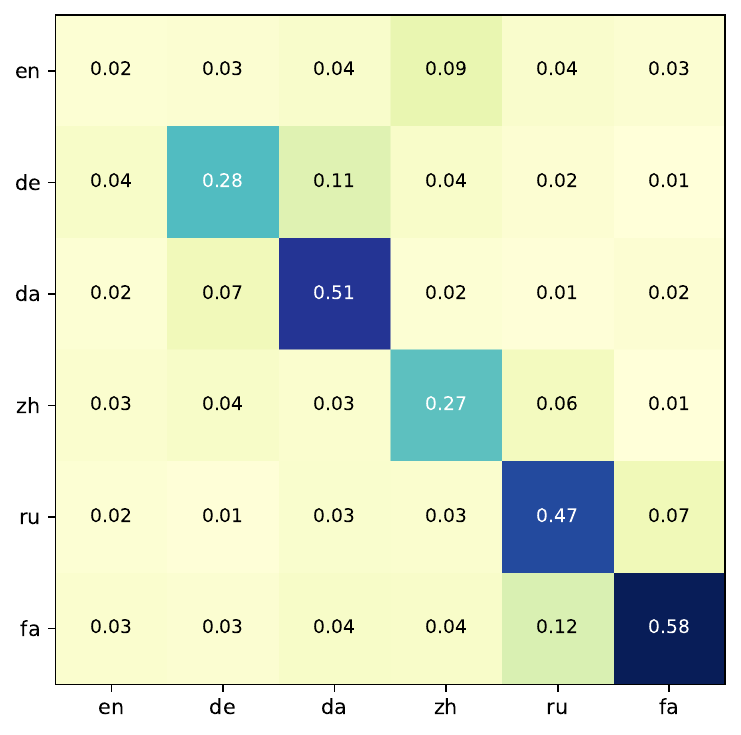}
\caption{Language $\cap$ Culture}
\end{subfigure}
\begin{subfigure}{0.49\linewidth}
\includegraphics[width=\linewidth]{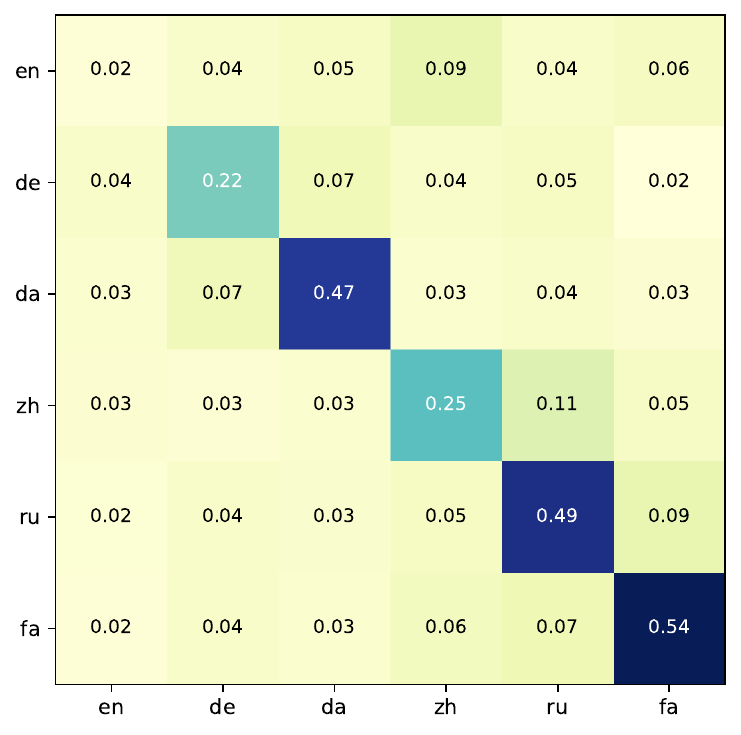}
\caption{Language}
\end{subfigure}\caption{Impact of ablating four neuron subsets on our MUREL test set in Llama-2-7b. Each cell $(i,j)$ shows perplexity (PPL) change on culture $j$ when ablating neurons of language or culture $i$.}
\label{fig:heatmap_Llama-2-7b}
\end{figure}
\begin{figure}[t!]
\centering
\begin{subfigure}{0.49\linewidth}
\includegraphics[width=\linewidth]{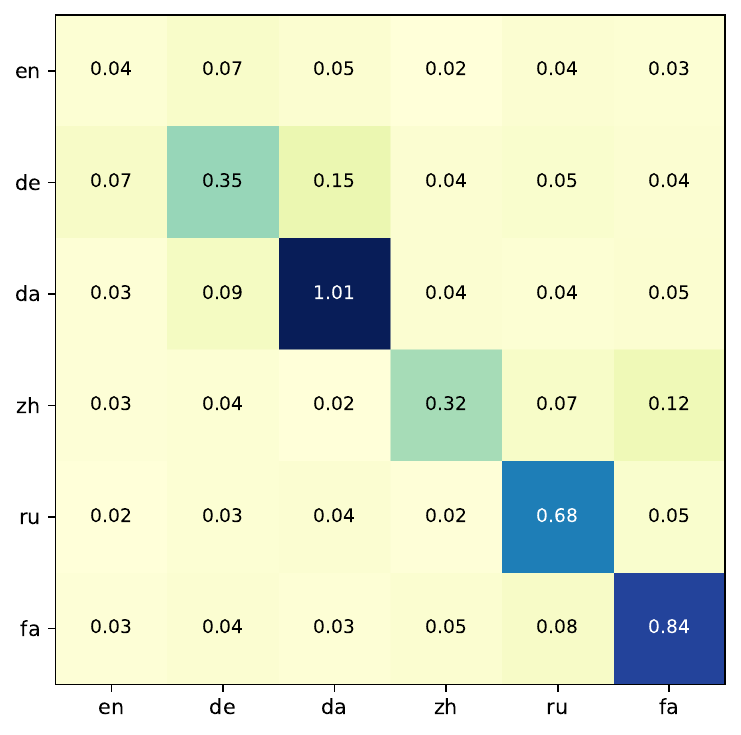}
\caption{Culture}
\end{subfigure}
\begin{subfigure}{0.49\linewidth}
\includegraphics[width=\linewidth]{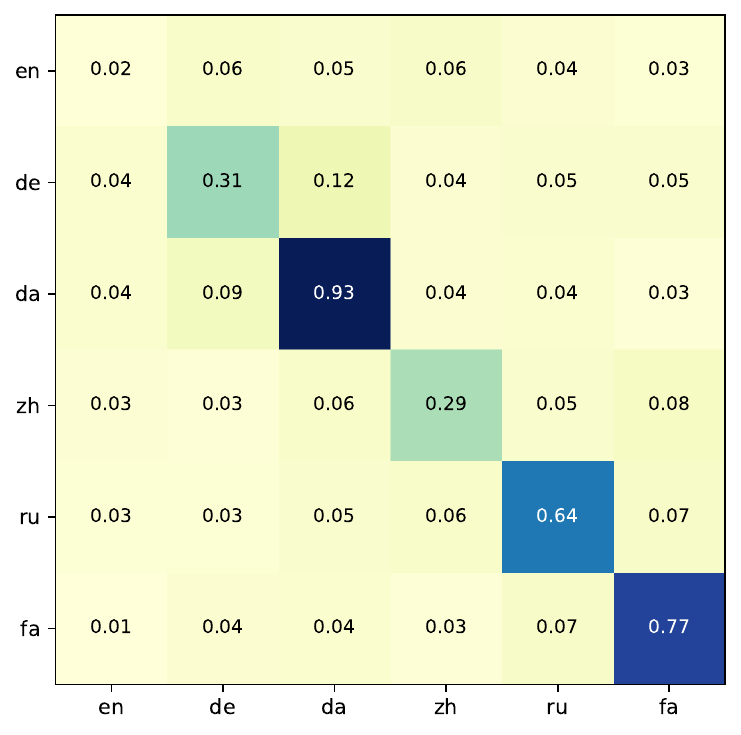}
\caption{Pure Culture}
\end{subfigure}
\begin{subfigure}{0.49\linewidth}
\includegraphics[width=\linewidth]{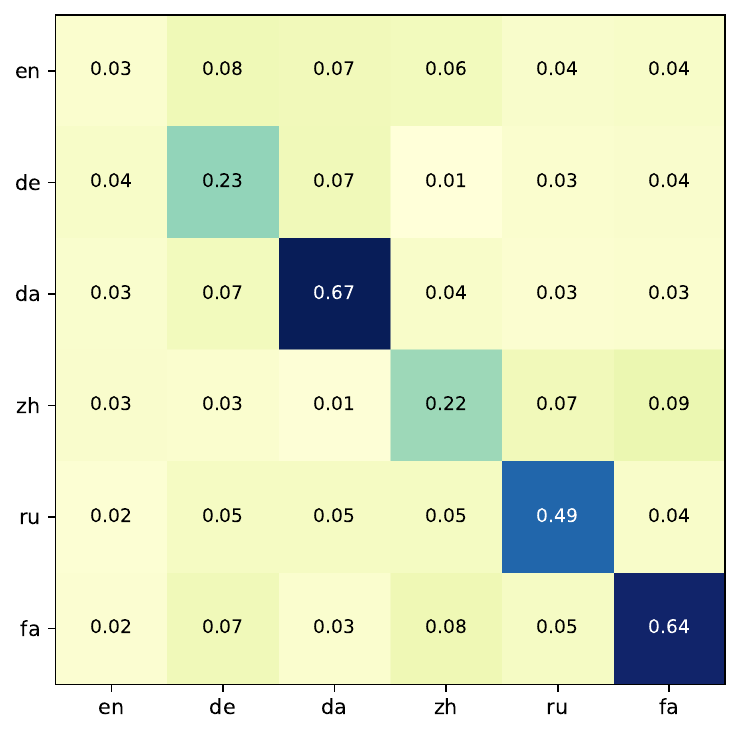}
\caption{Language $\cap$ Culture}
\end{subfigure}
\begin{subfigure}{0.49\linewidth}
\includegraphics[width=\linewidth]{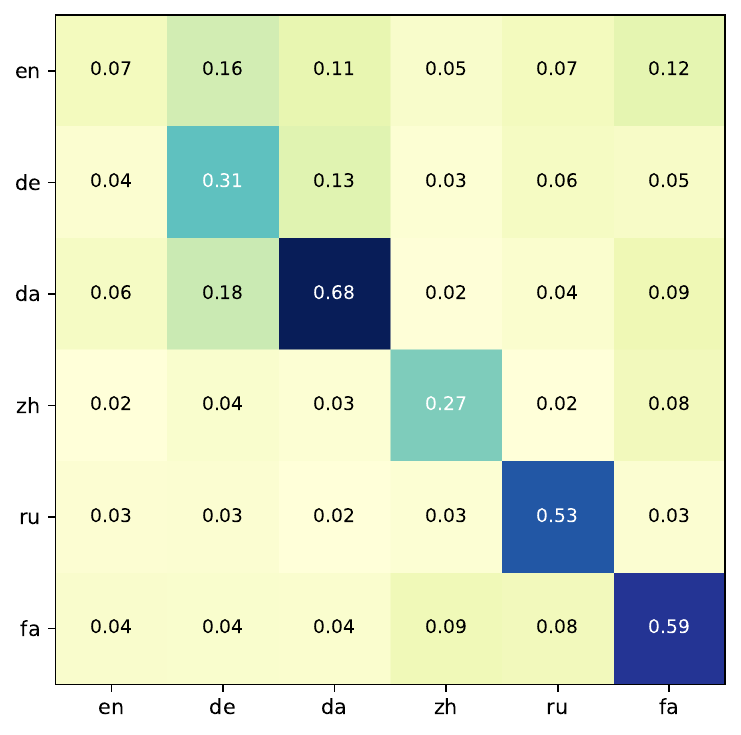}
\caption{Language}
\end{subfigure}\caption{Impact of ablating four neuron subsets on our MUREL test set in Llama-3.1-8b. Each cell $(i,j)$ shows the perplexity (PPL) change on culture $j$ when ablating neurons of language or culture $i$.}
\label{fig:heatmap_Llama-3.1-8b}
\end{figure}
\begin{figure}[t!]
\centering
\begin{subfigure}{0.49\linewidth}
\includegraphics[width=\linewidth]{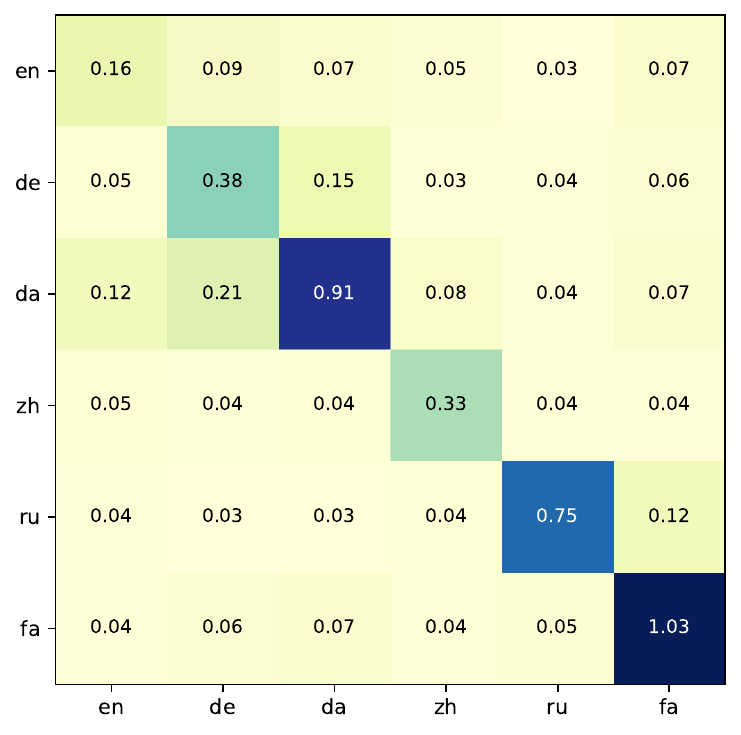}
\caption{Culture}
\end{subfigure}
\begin{subfigure}{0.49\linewidth}
\includegraphics[width=\linewidth]{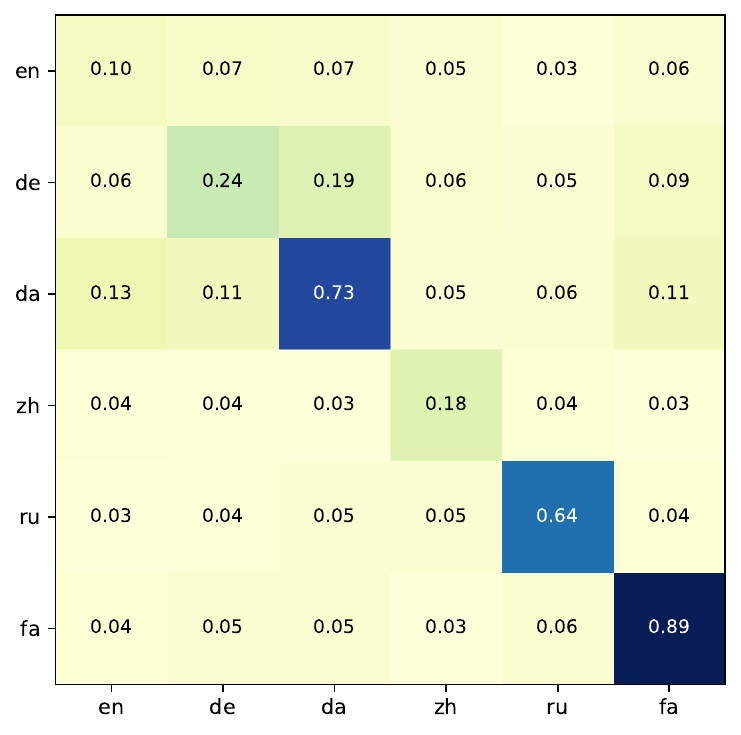}
\caption{Pure Culture}
\end{subfigure}
\begin{subfigure}{0.49\linewidth}
\includegraphics[width=\linewidth]{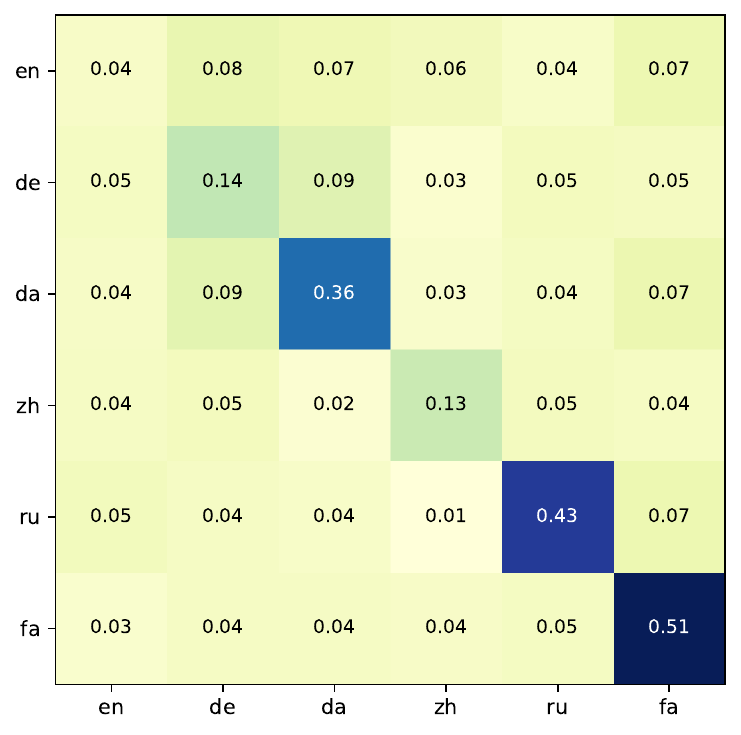}
\caption{Language $\cap$ Culture}
\end{subfigure}
\begin{subfigure}{0.49\linewidth}
\includegraphics[width=\linewidth]{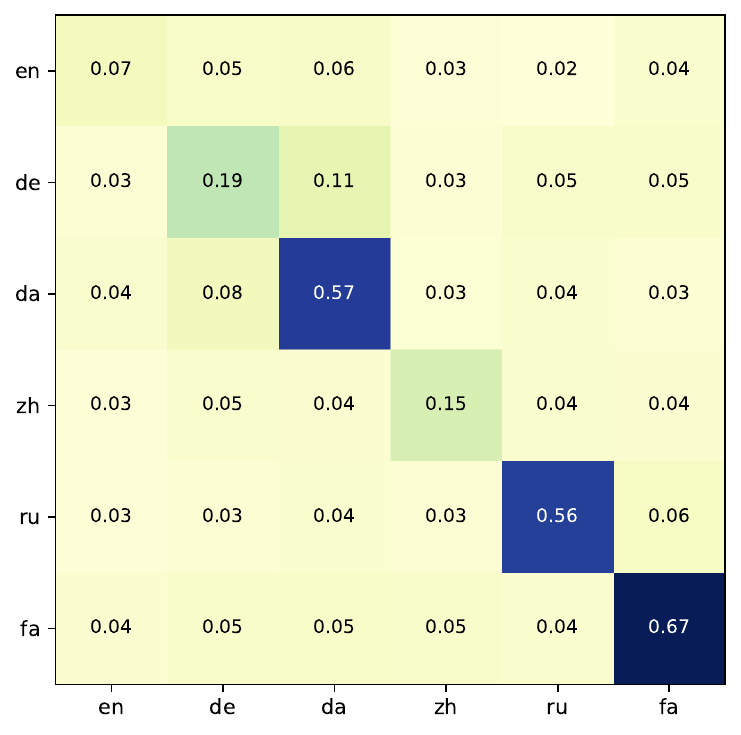}
\caption{Language}
\end{subfigure}\caption{Impact of ablating four neuron subsets on our MUREL test set in Qwen2.5-7b. Each cell $(i,j)$ shows perplexity (PPL) change on culture $j$ when ablating neurons of language or culture $i$.}
\label{fig:heatmap_Qwen2.5-7b}
\end{figure}
\begin{figure}[t!]
\centering
\begin{subfigure}{0.49\linewidth}
\includegraphics[width=\linewidth]{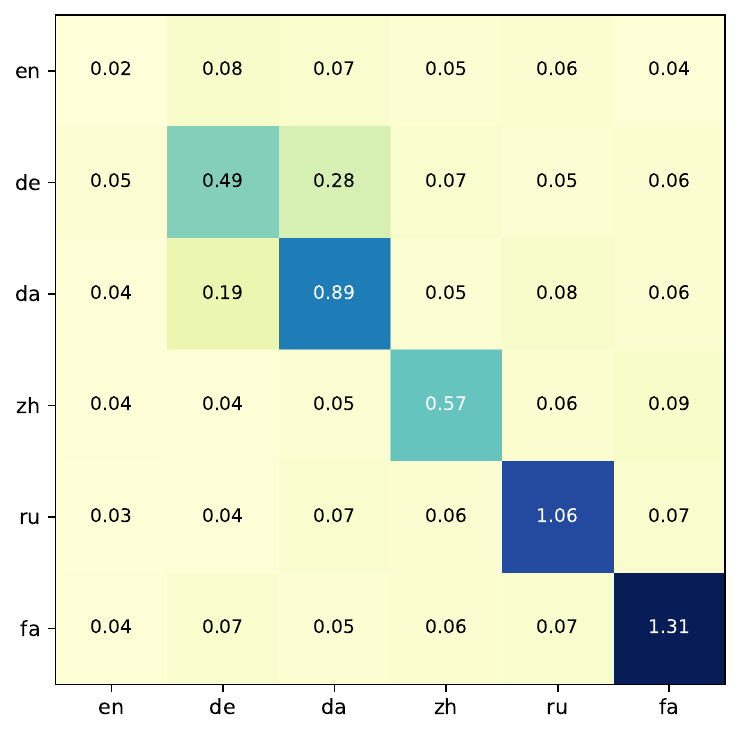}
\caption{Culture}
\end{subfigure}
\begin{subfigure}{0.49\linewidth}
\includegraphics[width=\linewidth]{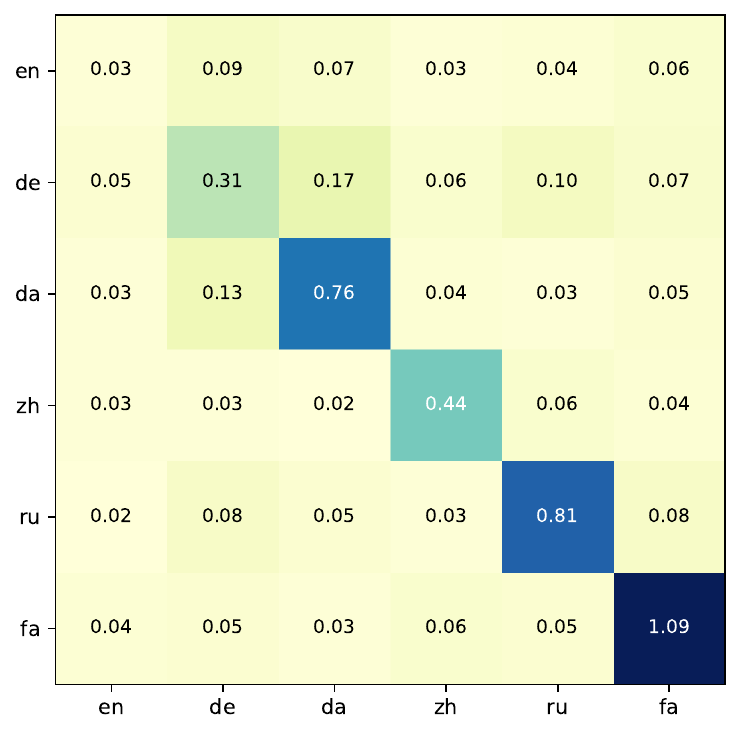}
\caption{Pure Culture}
\end{subfigure}
\begin{subfigure}{0.49\linewidth}
\includegraphics[width=\linewidth]{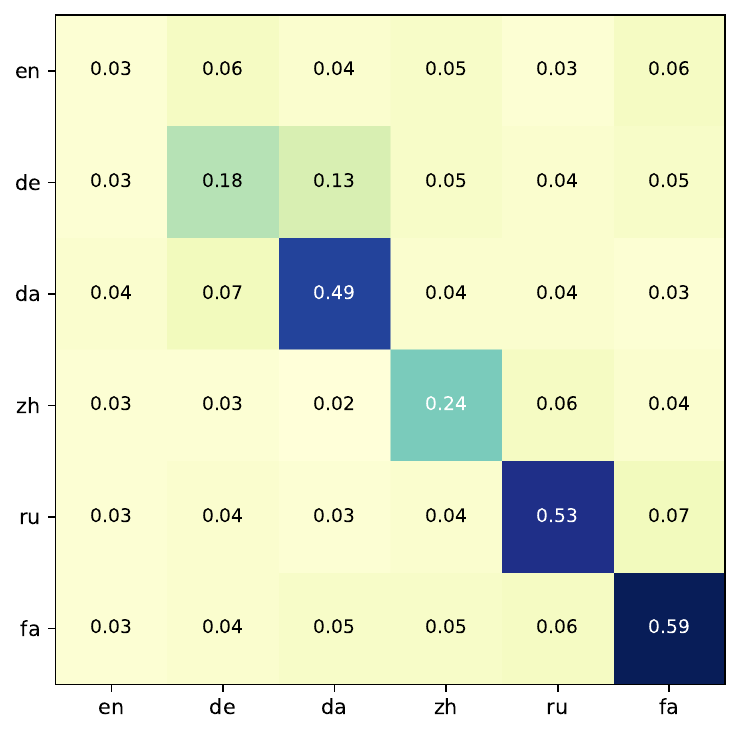}
\caption{Language $\cap$ Culture}
\end{subfigure}
\begin{subfigure}{0.49\linewidth}
\includegraphics[width=\linewidth]{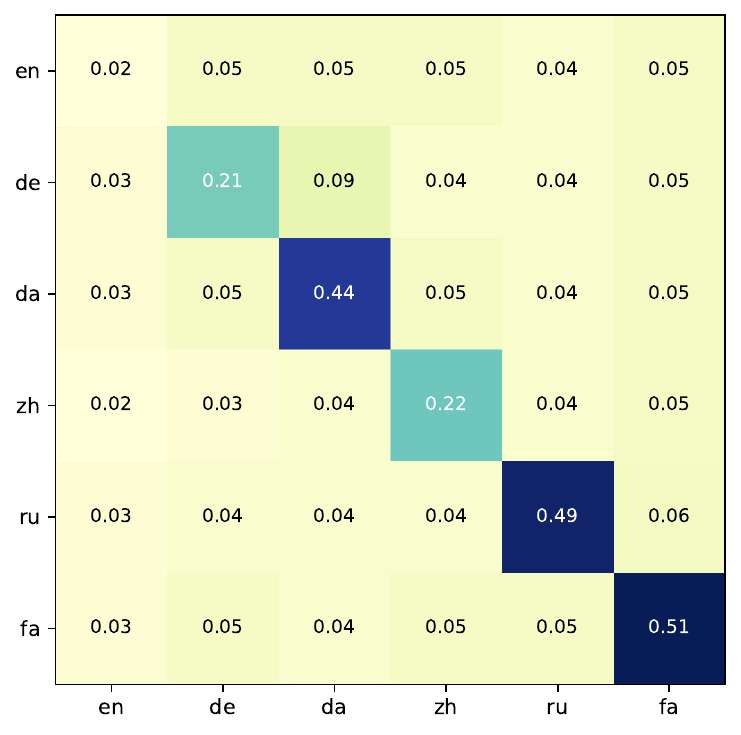}
\caption{Language}
\end{subfigure}\caption{Impact of ablating four neuron subsets on our MUREL test set in Gemma-3-12b. Each cell $(i,j)$ shows perplexity (PPL) change on culture $j$ when ablating neurons of language or culture $i$.}
\label{fig:heatmap_Gemma-3-12b}
\end{figure}

To assess the functional roles of identified neuron subpopulations, we conduct ablation experiments by zeroing out: (a) language-specific, (b) culture-specific, (c) pure culture-specific, (d) compound language-and-culture, and (e) randomly selected neurons. We then measure the resulting change in model perplexity on the MUREL dataset.

Figure~\ref{fig:heatmap_Llama-3.1-8b} illustrates perplexity changes in the Llama-3.1-8b model after these interventions; diagonal entries reflect effects within the corresponding language or culture, while off-diagonal entries indicate cross-linguistic or cross-cultural impact.

Ablating culture-specific neurons yields the largest increase in perplexity for culturally relevant data, confirming their critical role. Pure culture-specific neurons cause the second-largest perplexity increase, and accounts for about 76.3\% of the total effect from ablating all culture-specific neurons. This shows that a large share of cultural knowledge in LLMs is encoded in neurons that are largely independent of language processing.

\begin{figure}[!t]
    \centering
    \includegraphics[width=0.49\linewidth]{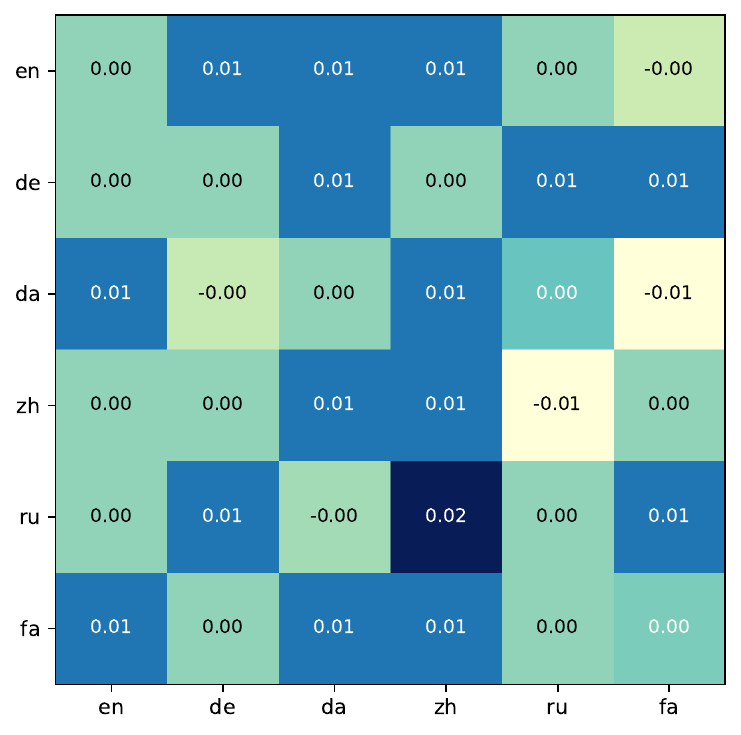}
    \caption{Perplexity changes after randomly ablating neurons in Llama-3.1-8b. Number of ablated neurons per culture matches the average identified per culture.}
    \label{fig:ppl_random}
\end{figure}

Crucially, off-diagonal (cross-linguistic and cross-cultural) effects remain consistently minimal, indicating that ablations mainly impact the targeted language or culture. Random neuron ablation has a negligible effect (Figure~\ref{fig:ppl_random}), further emphasizing that the functional roles of identified neuron groups are not due to chance. These patterns hold across all evaluated models, with only minor variation.

\section{Discussion and Conclusion}
We show that culture-specific neurons—and especially pure culture-specific neurons, which encode cultural knowledge independently of linguistic representations -- play a substantial role in shaping model predictions for culturally nuanced content. Although pure culture-specific neurons constitute only about 56.7\% of culture-related neurons, their ablation disproportionately increases perplexity, underscoring their functional importance. These results indicate that multilingual language models organize cultural knowledge into specialized neural populations, separable from linguistic encoding. Notably, both language- and culture-specific neurons predominantly reside in upper layers, consistent with hierarchical theories of semantic representation. Thus, our work enhances our understanding of how multilingual models internally represent complex cultural and semantic information.

This work advances our understanding of how cultural information is represented within multilingual language models. Our approach offers a framework for probing the interplay between cultural and linguistic signals in model internals, and facilitates future work on representational structure, identity modeling, or culturally grounded evaluation. Our findings suggest that, like language, culture can be meaningfully localized and examined as a distinct component of model representations.
Notably, we did not find any ``generic'' culture neurons shared across all cultures, i.e., $\bigcap_m \mathbb{P}_m = \emptyset$, indicating that cultural representations are highly specific.

\paragraph{Conclusion} We have introduced a methodology to identify and isolate culture neurons in multilingual language models.
To facilitate our analyses, we have compiled \textsc{MUREL}, a large and culturally diverse resource.
Our results show that cultural knowledge concentrates in specialized neuron populations predominantly localized in upper layers of multilingual language models and that pure culture neurons play a substantial functional role. Ablation experiments demonstrate that each culture is encoded in distinct neural populations with minimal cross-cultural interference.

We invite future research on (i)~broadening coverage to additional cultures, languages, and models, (ii)~extending localization to attention heads where the present study focuses on the feedforward modules, and (iii)~testing sufficiency via activation scaling and steering, as well as evaluating intervention effects on downstream tasks.

\section{Limitations}
While our study provides new insights into the neural localization of culture in multilingual language models, several limitations remain. First, our analysis is restricted to a small set of open-source models and may not generalize to larger or proprietary LLMs with different architectures or training data. Second, our methodology for identifying culture neurons relies on entropy-based metrics and dataset sampling choices, which may limit our ability to detect more distributed or context-dependent representations. Third, although the MUREL dataset is diverse, it covers only six cultures, potentially omitting important cultural phenomena found in other regions or language families. Finally, our evaluation focuses on neuron ablation and perplexity; future work should include more comprehensive behavioral and downstream assessments to better understand the practical impact of these neurons.

\section{Potential Risks}
This research analyzes the internal representations of multilingual language models and introduces MUREL, a culturally diverse evaluation dataset. All data used are derived from publicly available and properly credited resources. No private, sensitive, or personally identifiable information was included. While identifying culture-specific neurons may help increase transparency and cultural awareness in language models, it also raises the risk of model manipulation or the reinforcement of cultural stereotypes if misused. Our methodology is intended to advance understanding and fairness in multilingual NLP, not to entrench or amplify cultural biases.

\section*{Acknowledgements}
This work was supported in part by the Danish Foundation Models project.

\bibliography{yourbib}

% \clearpage

\tikzset{%
    parent/.style =          {align=center, text width=2cm, rounded corners=3pt, line width=0.3mm, fill=blue!80, draw=blue!80},
    child/.style =           {align=center, text width=2.3cm, rounded corners=3pt, fill=blue!70, draw=blue!80, line width=0.3mm},
    grandchild/.style =      {align=center, text width=2cm, rounded corners=3pt},
    greatgrandchild/.style = {align=center, text width=1.5cm, rounded corners=3pt},
    greatgrandchild2/.style = {align=center, text width=1.5cm, rounded corners=3pt},    
    referenceblock/.style =  {align=center, text width=1.5cm, rounded corners=2pt},
    data_pt/.style =         {align=center, text width=2cm, rounded corners=3pt, fill=paired-light-green!50, draw=paired-dark-green!65, line width=0.3mm},   
    data/.style =            {align=center, text width=2.5cm, rounded corners=3pt, fill=paired-light-green!50, draw=paired-dark-green!65, line width=0.3mm},   
    data_sub/.style =        {align=center, text width=1.5cm, rounded corners=3pt, fill=paired-light-green!50, draw=paired-dark-green!65, line width=0.3mm},   
    data_work/.style =       {align=left, text width=3.2cm, rounded corners=3pt, fill=paired-light-green!50, draw=paired-dark-green!80, line width=0.3mm} 
}

\begin{figure*}[]
    \scriptsize
    \centering
    \resizebox{0.9\textwidth}{!}{
    \begin{forest}
        for tree={
            forked edges,
            grow'=0,
            draw,
            rounded corners,
            node options={align=center,},
            text width=2cm,
            s sep=4pt,
            calign=child edge, 
            calign child=(n_children()+1)/2,
            l sep=7.5pt,
        },
        [, phantom
            [Data \S\ref{sec:datasets}, data_pt
                [Ideational, 
                data[Concepts, data_sub[
                                % \textbf{Metaphor:} 
                                \citet{mohler-etal-2016-introducing, schneider-etal-2022-metaphor, kabra-etal-2023-multi, piccirilli-etal-2024-volimet, kim2023metaphorian}
                                % \textbf{Idiom:} 
                                \citet{aharodnik-etal-2018-designing, moussallem-etal-2018-lidioms, adewumi-etal-2022-potential, saxena2020epie, stap-etal-2024-fine, khoshtab-etal-2025-comparative, rezaeimanesh-etal-2025-large, sorensen-nimb-2025-danish}
                                % \textbf{Simile:}
                                \citet{chakrabarty-etal-2022-rocket, chakrabarty-etal-2022-flute}
                                % \textbf{Irony:}
                                \citet{golazizian-etal-2020-irony}
                                % \citet{casola-etal-2024-multipico}
                                % \textbf{Proverb:} 
                                \citet{liu-etal-2024-multilingual}, data_work]]
                data[Knowledge, data_sub[
                \citet{koto-etal-2024-indoculture, shi-etal-2024-culturebank, bhatt-diaz-2024-extrinsic, wang-etal-2024-seaeval, fung2024massively, zhou-etal-2025-mapo}, data_work]]
                data[Values, data_sub[
                \citet{mohamed-etal-2022-artelingo, cao-etal-2023-assessing, pistilli2024civics, lee-etal-2024-exploring-cross, attanasio-etal-2023-tale, abdelkadir-etal-2024-diverse, casola-etal-2024-multipico, sap-etal-2022-annotators}, data_work]]
                data[{Norms and Morals}, data_sub[
                \citet{sun-etal-2023-moraldial, li-etal-2023-normdial, dwivedi-etal-2023-eticor, ch-wang-etal-2023-sociocultural, fung-etal-2023-normsage, yuan-etal-2024-measuring, yu-etal-2024-cmoraleval}, data_work] ]
                data[Artifacts, data_sub[
                \citet{yang-etal-2019-generating, chakrabarty-etal-2021-dont, schmidt-etal-2021-fairynet, ou-etal-2023-songs, khashabi-etal-2021-parsinlu}, data_work] ]
                ]
                [Linguistic,  
                data[Dialects, data_sub[\citet{malmasi-zampieri-2017-german, ciobanu-etal-2018-german, pluss-etal-2023-stt4sg, kuparinen-etal-2023-dialect, paonessa-etal-2023-dialect, abaskohi-etal-2024-benchmarking}, data_work]]
                data[{Styles, Registers, Genres}, data_sub[\citet{sun-xu-2022-tracing, nadejde-etal-2022-cocoa, srinivasan-choi-2022-tydip, havaldar-etal-2023-comparing}, data_work]]
                ]
                [Social,
                data[Relationship, data_sub[\citet{}\\
                \citet{abs-2402-11178}, data_work]]
                data[Context, data_sub[
                \citet{hovy-etal-2020-sound, chakrabarty-etal-2022-rocket, zhan2023social, ziems-etal-2023-normbank}, data_work]]
                data[Communicative Goals, data_sub[
                \citet{emelin-etal-2021-moral, li-etal-2023-normdial, abs-2402-11178}, data_work]]
                data[Demographics, data_sub[
                \citet{hovy-2015-demographic, voigt-etal-2018-rtgender, hovy-etal-2020-sound, pei-jurgens-2023-annotator, santy-etal-2023-nlpositionality, frenda-etal-2023-epic, abs-2308-16705}, data_work]]
                ]
            ]
        ]
    \end{forest}
}
    \caption{Categorization of cultural data resources in MUREL, with representative references for each category.}
    \label{fig:overviews_data}
\end{figure*}
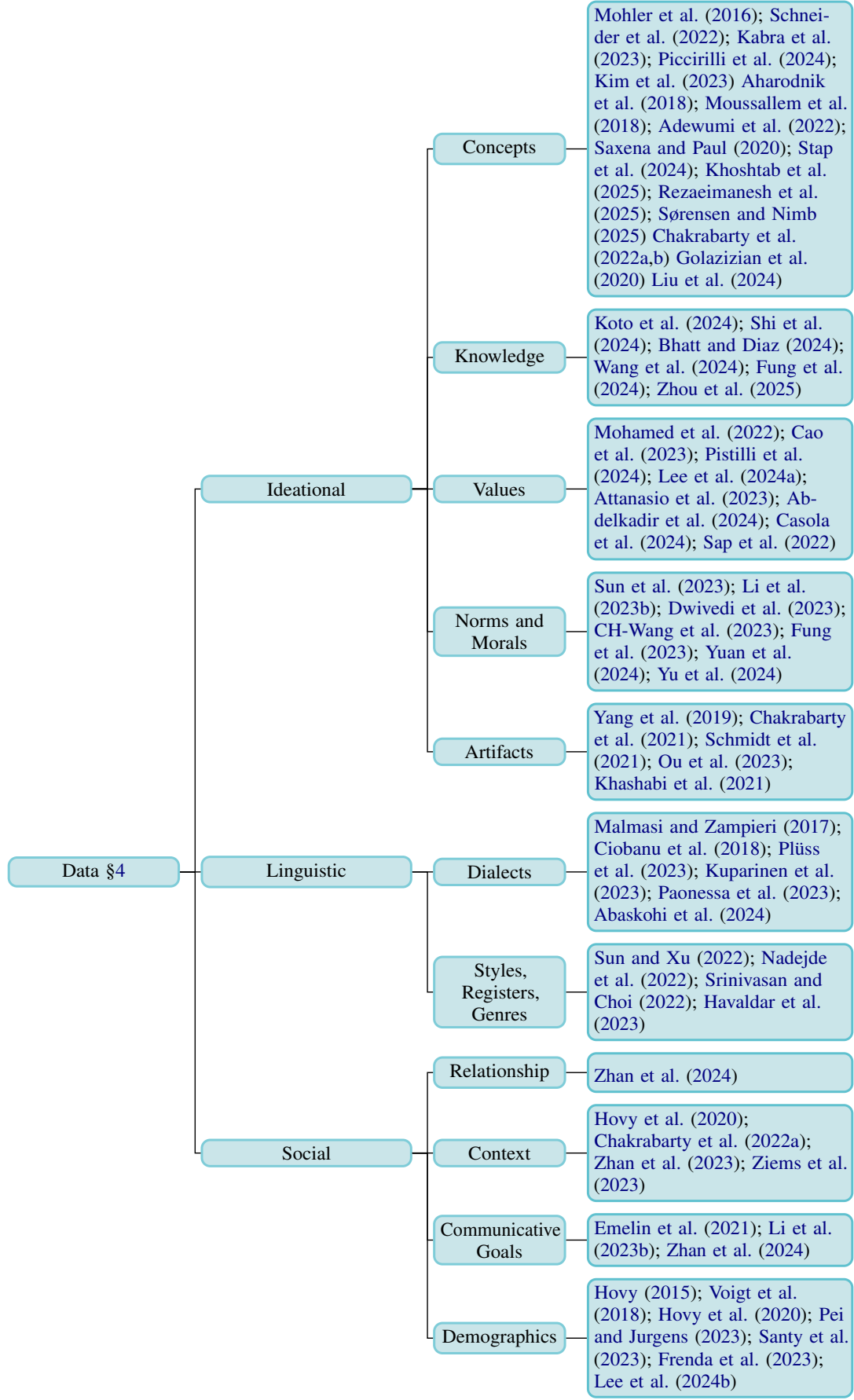

\appendix
\section{Datasets}\label{app:A}
\label{app:datasets}

We compiled the MUREL dataset, consisting of 69 culturally diverse corpora spanning 6 cultures, with a total of 85.2 million Gemma-3 tokens.
% For each culture, we aimed for broad coverage across ideational, linguistic, and social domains.
All datasets used in this study are publicly available and were used in accordance with their respective open licenses for research purposes only.
Table~\ref{tab:dataset_stats} reports the total number of tokens per culture.
Figure~\ref{fig:overviews_data} shows the systematic organization and provides references to all source datasets.

\begin{table}[ht]
  \centering
  \resizebox{1.0\columnwidth}{!}{
    \begin{tabular}{cccccc}
    \toprule
    \textbf{EN} & \textbf{DE} & \textbf{DA} & \textbf{ZH} & \textbf{RU} & \textbf{FA}\\
    \midrule
    14,262 & 19,891 & 11,383 & 16,511 & 12,405 & 10,769\\
    \bottomrule
    \end{tabular}}
  \caption{Total number of tokens per culture in MUREL (in thousands).}
  \label{tab:dataset_stats}
\end{table}

\section{Models}\label{app:B}
For our investigation, we select four transformer-based language models.

\subsection{Llama 2}
Llama 2\footnote{\url{https://huggingface.co/meta-llama/Llama-2-7b}} is a 7-billion-parameter decoder-only transformer model developed by Meta. It consists of 32 layers and 352,256 neurons. It was trained on a corpus comprising approximately 2 trillion tokens of publicly available online data. While Llama 2 supports text generation in English and 27 other languages, its training data is predominantly English, which may affect performance in less-represented languages.

\subsection{Llama 3.1}
Llama 3.1\footnote{\url{https://huggingface.co/meta-llama/Llama-3.1-8B}} is an 8-billion-parameter multilingual model from Meta, with 32 layers and 458,752 neurons. It was trained on diverse text corpora. The model consists of stacked transformer layers, each comprising self-attention and feedforward MLP components. Llama 3.1 is optimized for computational efficiency and supports a wide range of languages, making it a strong candidate for evaluating multilingual transfer performance.

\subsection{Gemma 3}
Gemma 3\footnote{\url{https://huggingface.co/google/gemma-3-12b-pt}} is a 12-billion-parameter transformer-based model developed by Google, with 48 layers and 737,280 neurons. It is part of the Gemma family of lightweight, open models built from the same research and technology used to create Gemini. Gemma 3 models are multimodal and have a large, 128K context window, multilingual support in over 140 languages, and are available in more sizes than previous versions.

\subsection{Qwen 2.5}
Qwen 2.5\footnote{\url{https://huggingface.co/Qwen/Qwen2.5-7B}} is a 7-billion-parameter decoder-only transformer model developed by Alibaba Cloud, comprising 28 layers and 100,352 neurons. The Qwen2.5-7B model was trained on a substantial corpus of 18 trillion tokens, significantly expanding upon the 7 trillion tokens used in its predecessor, Qwen2. The model supports over 29 languages, making it a robust choice for multilingual applications.

\begin{figure}[!ht]
    \centering
    \includegraphics[width=0.49\linewidth]{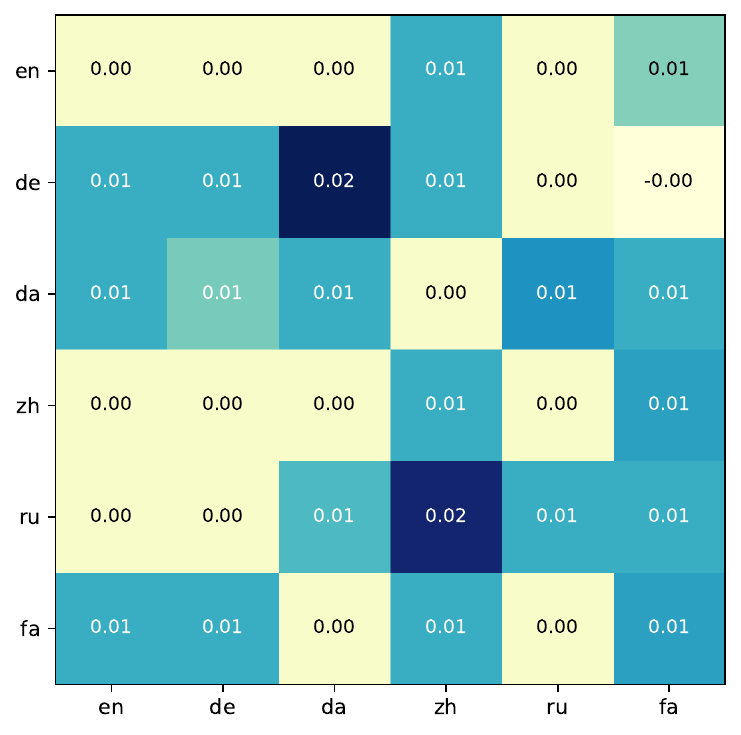}
    \caption{Perplexity changes after randomly ablating neurons in Llama-2-7b. Number of ablated neurons per culture matches the average identified per culture.}
    \label{fig:llama-2_ppl_random}
\end{figure}

\begin{figure}[!ht]
    \centering
    \includegraphics[width=0.49\linewidth]{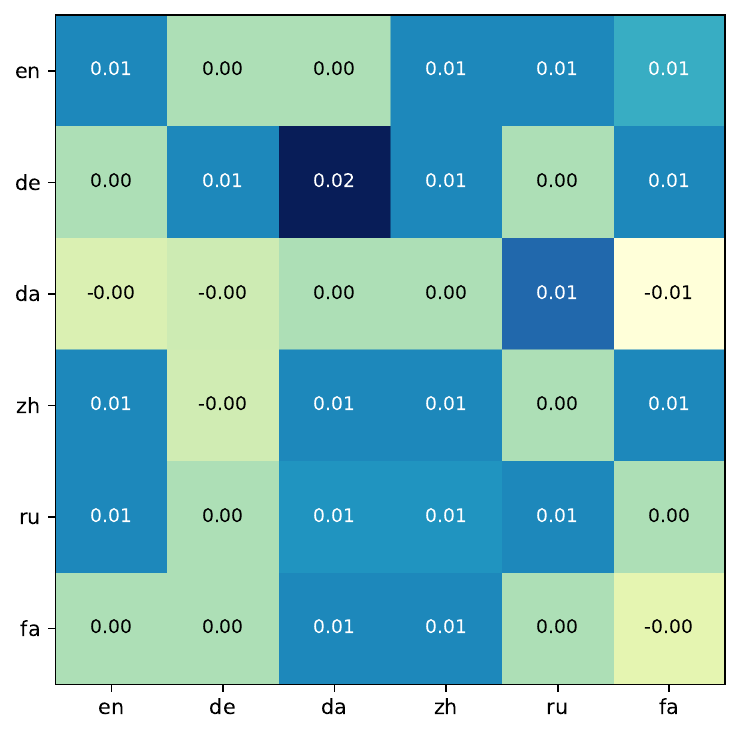}
    \caption{Perplexity changes after randomly ablating neurons in Qwen2.5-7b. Number of ablated neurons per culture matches the average identified per culture.}
    \label{fig:qwen2.5_ppl_random}
\end{figure}

\begin{figure}[!ht]
    \centering
    \includegraphics[width=0.49\linewidth]{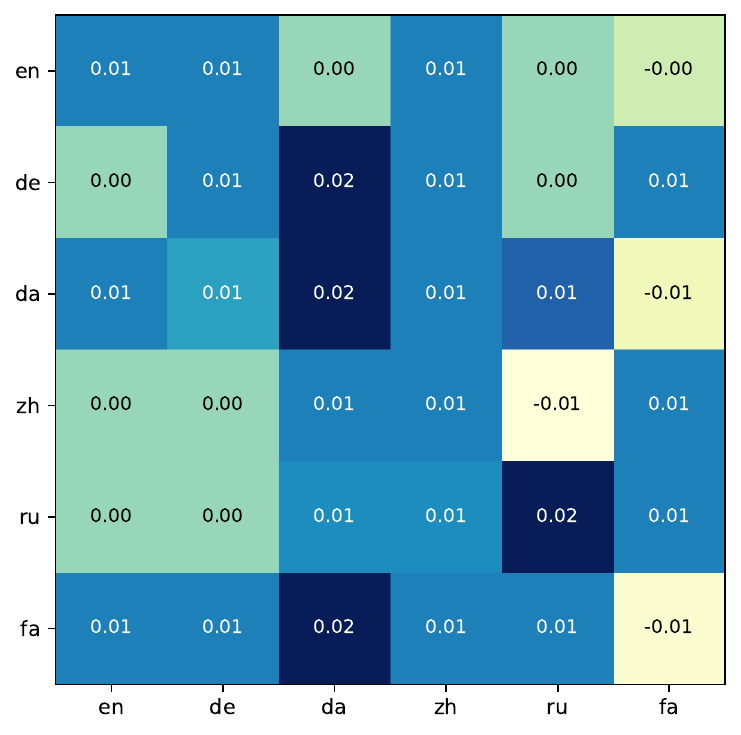}
    \caption{Perplexity changes after randomly ablating neurons in Gemma-3-12b. Number of ablated neurons per culture matches the average identified per culture.}
    \label{fig:gemma-3_ppl_random}
\end{figure}

\section{Additional Random Ablations}
Figures~\ref{fig:llama-2_ppl_random},  \ref{fig:qwen2.5_ppl_random}, and \ref{fig:gemma-3_ppl_random} show the random ablations for Llama-2-7b, Qwen2.5-7b, and Gemma-3-12b, respectively.

\section{Computational Infrastructure}
All experiments, including neuron activation analysis and ablation interventions, were conducted using pretrained models without any additional training or fine-tuning. Computations were performed on a single NVIDIA V100 GPU per experiment. Across all models, the total computational budget did not exceed 280 GPU hours.

\begin{figure*}[p]  % Use [p] for float page, or [htbp] as needed
    \centering
    \begin{subfigure}[b]{0.95\textwidth}
        \centering
        \includegraphics[width=\textwidth]{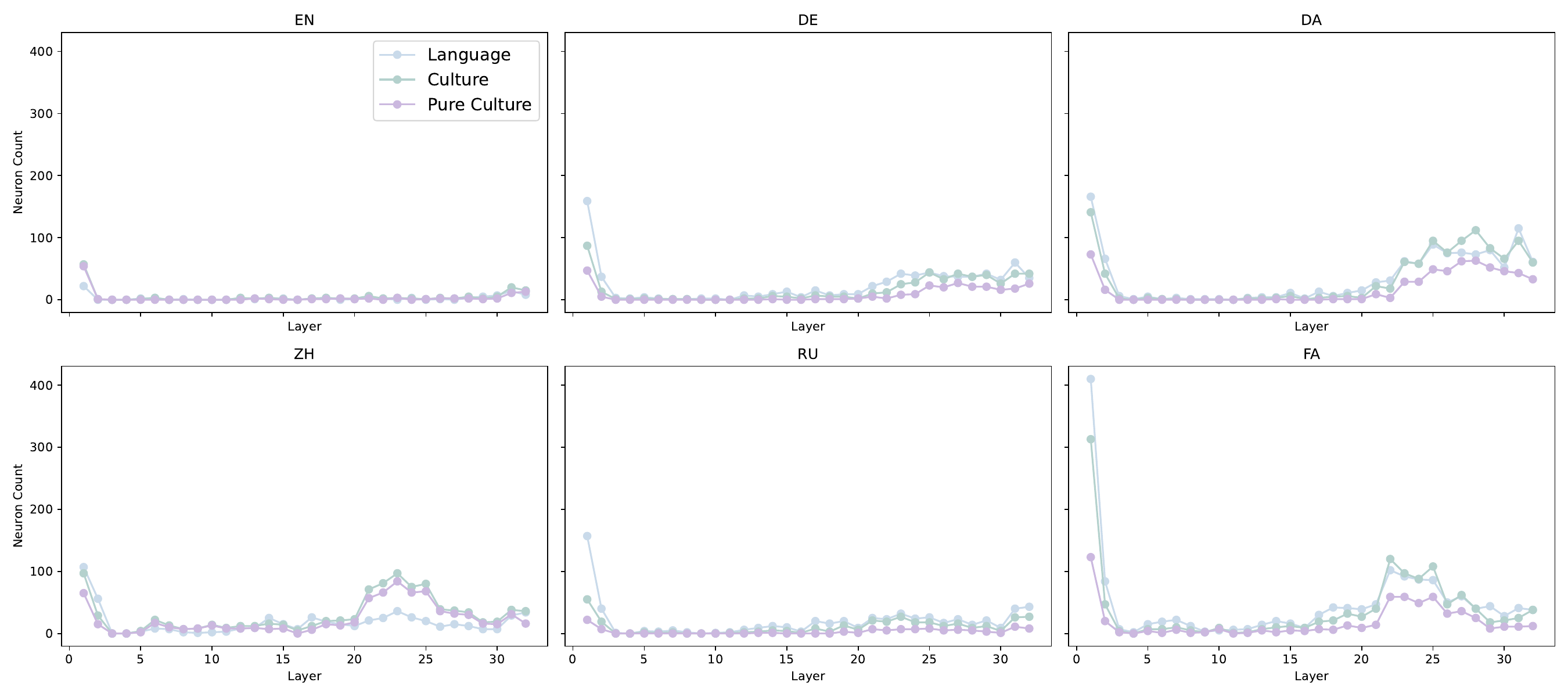}
        \caption{Llama-2-7b}
        \label{fig:lineplot_Llama-2-7b}
    \end{subfigure}
    \vspace{4mm}
    \begin{subfigure}[b]{0.95\textwidth}
        \centering
        \includegraphics[width=\textwidth]{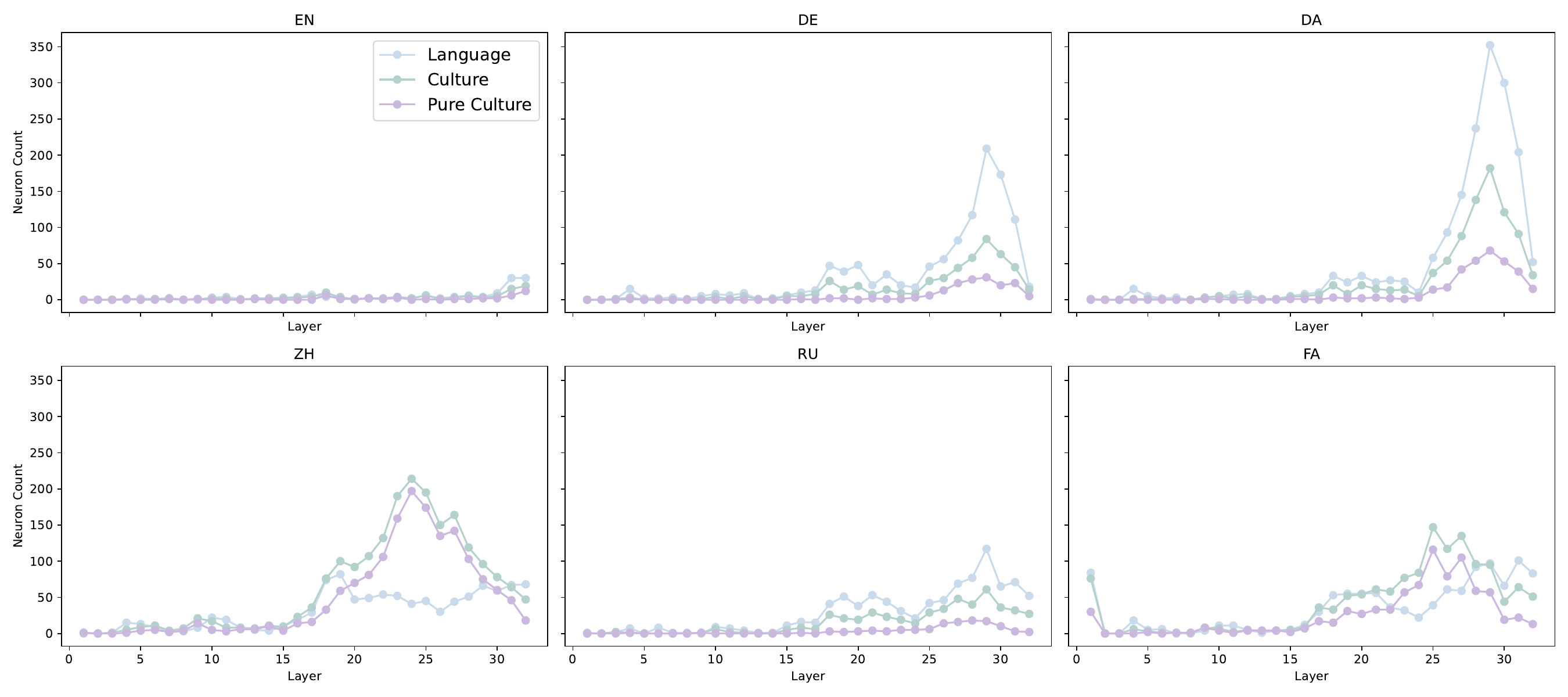}
        \caption{Llama-3.1-8b}
        \label{fig:lineplot_Llama-3.1-8b}
    \end{subfigure}
    \caption{
        Layer-wise distribution of language, culture, and pure culture neurons for each language, 
        visualized for (a) Llama-2-7b and (b) Llama-3.1-8b.
    }
    \label{fig:all_languages_layerwise_line_both}
\end{figure*}

% \bibliographystyle{acl_natbib}
% \footnotesize

\end{document}